\documentclass[11pt]{article}
\usepackage{acl}

\usepackage{times}
\usepackage{latexsym}
\usepackage[T1]{fontenc}
\usepackage[utf8]{inputenc}
\usepackage{microtype}
\usepackage{inconsolata}
\usepackage{graphicx}

\usepackage{amsmath,amssymb,mathtools,amsthm}
\usepackage{booktabs,tabularx,multirow,array}
\usepackage{xcolor,pifont}
\usepackage[most]{tcolorbox}
\usepackage{caption}
\usepackage{listings}
\usepackage{enumitem}
\usepackage{subfig}
\usepackage{algorithm}
\usepackage{algorithmic}
\usepackage[capitalize,noabbrev]{cleveref}

\theoremstyle{plain}

\theoremstyle{definition}

\theoremstyle{remark}

\newcommand{\benchname}{\textsc{AsyncTool}}
\newcommand{\cmark}{\textcolor{green!80!white}{\ding{51}}}
\newcommand{\xmark}{\textcolor{red}{\ding{55}}}
\newcommand{\eg}{\emph{e.g.}}

\title{AsyncTool: Evaluating the Asynchronous Function Calling Capability under Multi-Task Scenarios}

\author{
\textbf{Kou Shi}\textsuperscript{1\textdagger},
\textbf{Ziao Zhang}\textsuperscript{1\textdagger},
\textbf{Shiting Huang}\textsuperscript{1},
\textbf{Avery Nie}\textsuperscript{2},
\textbf{Zhen Fang}\textsuperscript{1},\\
\textbf{Qiuchen Wang}\textsuperscript{1},
\textbf{Lin Chen}\textsuperscript{1},
\textbf{Huaian Chen}\textsuperscript{1},
\textbf{Zehui Chen}\textsuperscript{1},
\textbf{Feng Zhao}\textsuperscript{1*}\\
\textsuperscript{1}MoE Key Lab of BIPC, University of Science and Technology of China\\
\textsuperscript{2}University of Toronto\\
Contact: \texttt{stokou@mail.ustc.edu.cn}
\quad
\texttt{fzhao956@ustc.edu.cn}
}

\begin{document}
\maketitle
\begingroup
\renewcommand{\thefootnote}{}
\footnotetext{\textsuperscript{\textdagger}Equal contribution. \textsuperscript{*}Corresponding author.}
\addtocounter{footnote}{-1}
\endgroup

\begin{abstract}
    Large language model (LLM)-based agents have demonstrated strong capabilities in leveraging external tools to solve complex tasks. However, existing evaluations largely overlook the temporal dimension of tool invocation, particularly the impact of tool response latency, and are typically limited to single-task settings. In real-world applications, multiple tasks often need to be executed concurrently, and overall efficiency critically depends on whether an agent can utilize idle time while waiting for tool responses. We refer to this capability as asynchronous tool calling. To evaluate this capability, we propose AsyncTool, a benchmark for assessing LLM-based agents in interactive multi-task tool-use environments with delayed tool feedback. AsyncTool presents multiple heterogeneous tasks simultaneously and simulates realistic tool response latency during execution. Using a hybrid data evolution strategy, we construct a diverse asynchronous multitasking dataset covering multiple scenarios and tool-use patterns. We evaluate models at three levels—step, sub-task, and task—and further introduce efficiency-oriented metrics to measure task coordination and completion efficiency. Extensive experiments show that delayed tool feedback poses substantial challenges to current agents, leading to clear performance degradation. Models that better coordinate task switching and dependency tracking tend to achieve stronger performance on AsyncTool. Our analysis identifies the main failure modes of current tool agents and provides practical guidelines for designing future systems with stronger temporal reasoning and coordination capabilities. The code is available at https://github.com/StoKou/repo-asynctool.
\end{abstract}
\section{Introduction}
\label{sec:introduction}

Recent advances in large language models (LLMs) have significantly improved their ability to follow instructions and understand context, leading to increasingly capable LLM-based agents for tool use\citep{openai2025o3o4mini,comanici2025gemini,anthropic2025claude4,yang2025qwen3,team2025kimi,zeng2025glm,chen2025minimax,wang2025vrag,chen2024mindsearch,huang2026internalizing,zhang2026skillflow}. This capability enables them to handle more sophisticated, multi-step tasks that require external information or actions, and to achieve strong performance across diverse tool-use scenarios \citep{liu2023agentbench, li2025tool, chan2024mle}.

However, real-world environments are often more complex, frequently requiring the concurrent execution of multiple tasks that may involve different tools. In practical settings, function calls usually incur latency, and executing tasks sequentially in a synchronous manner fails to fully utilize idle waiting time, thereby reducing overall efficiency. To better evaluate and enhance the agent's performance under such conditions, we introduce the concept of \textbf{Asynchronous Tool Call} into the interaction between the agent and the environment, where the agent should utilize these idle intervals to advance other available tasks.
Motivated by these gaps, we identify three critical observations: \textbf{(i) Inadequate evaluation of the agent's capability to complete multiple tasks in asynchronous scenarios.} Existing studies are typically restricted to single-task scenarios in which tools operate in an immediate response manner \citep{zhuang2023toolqa, ruan2023tptu, xu2023tool, guo-etal-2024-ctooleval, qin2023toolllm, ye2024tooleyes}, overlooking the evaluation for multiple tasks in asynchronous scenarios.
\textbf{(ii) Lack of alignment with real-world conditions in interactive environments involving real-time tool calls.} Existing asynchronous planning benchmarks do not operate within interactive environments, which is inconsistent with real-world scenarios involving real-time tool calls \citep{10.5555/3692070.3693283}.
\textbf{(iii) Insufficient metrics and standardized protocols specific to concurrent tasks with delayed and out-of-order tool feedback.} Traditional benchmarks involving time delays do not cover tool-using tasks and cannot be transferred to agentic tasks \citep{zhang2024timearena,gonzalez2025robotouille}.

To bridge these gaps, we propose \benchname, a benchmark for evaluating the ability of LLM-based agents to perform asynchronous tool calling in interactive multi-task scenarios. To our knowledge, {\benchname} is the first benchmark that jointly considers delayed tool feedback, concurrent multi-task execution, multi-step function calling, and dependency-aware task coordination. 
(i) Our benchmark consists of combinations of multiple tasks, where each task contains intra-task step dependencies and different tasks can be pursued concurrently. This design allows an agent to use the waiting periods caused by tool latency to advance other independent tasks. 
(ii) To better approximate real-world tool-use conditions, we simulate tool-specific response latencies, integrate multiple tasks into a shared interaction process, and require the agent to make progress on them through asynchronous function calls. This setting provides a practical environment for assessing whether agents can coordinate multiple tasks under delayed and potentially out-of-order tool feedback. Table~\ref{table:compare_benchmarks} compares {\benchname} with existing benchmarks on tool calling and asynchronous execution. 
(iii) To comprehensively evaluate asynchronous tool-use capabilities, we assess model performance at three levels: Step Level, Sub-Task Level, and Task Level, covering fine-grained tool-call correctness, intermediate subtask completion, and end-to-end multi-task success. In addition, we introduce efficiency-oriented metrics to measure task-interleaving behavior and completion efficiency under tool latency.
Through extensive experiments, we find that delayed tool feedback poses substantial challenges to current LLM-based agents, especially in maintaining task states before results arrive. Compared with synchronous or immediate-response settings, asynchronous execution leads to clear performance degradation, especially when models prematurely continue a task before its dependent tool result has returned. Our analysis further shows that effective asynchronous tool use requires more than frequent task switching: models must coordinate task switching with dependency tracking and state maintenance. Stronger models are better able to utilize idle waiting periods to advance other tasks while resuming pending tasks at the appropriate time, whereas weaker models often suffer from dependency violations, task neglect, and tool confusion. These findings highlight the importance of temporal coordination for future tool-using agents.

The main contributions of our work are summarized as follows:
\begin{itemize}[]
\item
We propose \benchname, a benchmark for evaluating asynchronous tool calling in interactive multi-task environments with delayed tool feedback.
\item 
We construct a diverse asynchronous multitasking dataset by composing validated single-task tool-use trajectories through a hybrid data-evolution strategy. The resulting tasks cover different task numbers, task types, scenarios, and dependency structures.

\item  
We design a multi-level evaluation protocol that assesses model performance at the step, sub-task, and task levels, capturing both fine-grained tool-call correctness and end-to-end task completion.

\item 
We introduce efficiency-oriented metrics to analyze task interleaving and completion behavior under tool latency, and conduct extensive experiments to reveal the challenges current LLM agents face in temporal coordination and dependency tracking.
\end{itemize}

\begin{figure*}[!t]
    \centering
    \includegraphics[width=\textwidth]{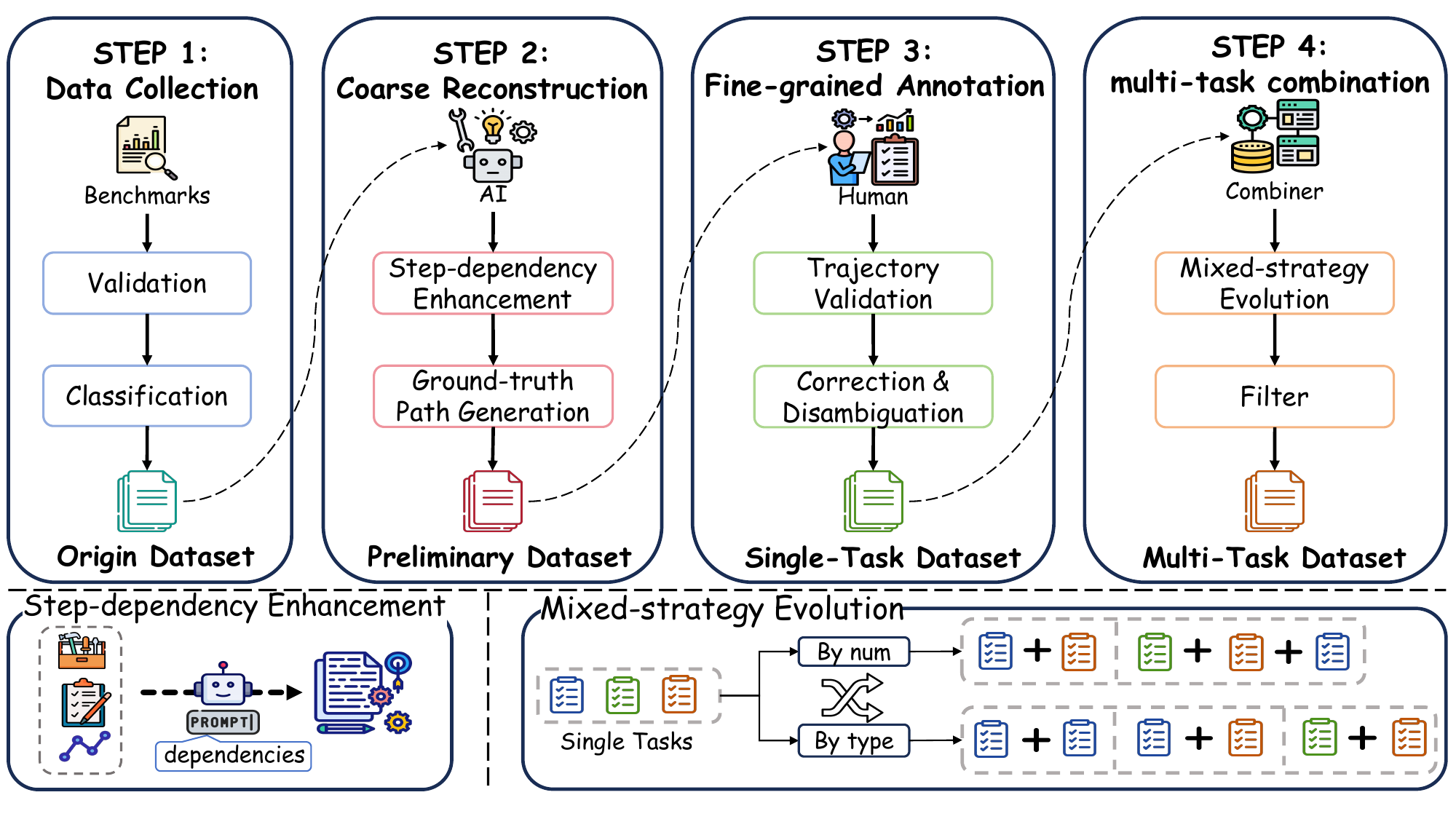}
    \caption{\label{fig:data_construction} 
    \textbf{Overview of the dataset construction process.} The pipeline starts by collecting raw data from tool-use benchmarks, categorizing it by scenario. Task-step dependencies are reinforced, and execution trajectories are reconstructed using Gemini 2.5 Pro, followed by manual verification for accuracy and determinism. Finally, data evolution occurs through hybrid strategies, with filtering producing the final multitasking dataset.}
\end{figure*}

\section{\benchname}

Building on the motivation introduced in Section~\ref{sec:introduction}, {\benchname} simulates tool response latency and evaluates asynchronous tool calling in multi-task settings. This section first formalizes the interaction paradigm in {\benchname}, where agents must coordinate multiple tasks under delayed tool feedback. We then describe the asynchronous multitasking dataset and evaluation protocol.

\subsection{Agent as a Concurrent Tool-Using System}

While agents' ability to solve problems through tool use is well established, their practical effectiveness can be limited by the non-negligible response latencies of tool calls in real-world scenarios.

This raises an important question: can agents use idle time from pending tool calls to work on other tasks? {\benchname} studies this by simulating delayed tool feedback and concurrent execution. Unlike standard tool-use settings, tool results in {\benchname} are returned with delays. After making a tool call, the agent must decide whether to wait or switch to another task. This makes delayed feedback, task interleaving, and dependency-aware scheduling key challenges in {\benchname}.

For example, consider a scenario where the agent receives two independent tasks, denoted as $task_1$ and $task_2$, whose required function-call sequences are $\langle f_1, f_3, f_5\rangle$ and $\langle f_2, f_4\rangle$, respectively. Although the two tasks are mutually independent, the function calls within each task must follow the specified order due to intra-task dependencies. In this setting, the agent acts as the \textit{Assistant}, while the execution system serves as the \textit{Environment}. The \textit{Assistant} first attempts to solve $task_1$ by calling $f_1$. After receiving the formatted tool-call request, the \textit{Environment} informs the \textit{Assistant} that the result of $f_1$ is not yet available, since tool execution is non-instantaneous. The \textit{Assistant} can then switch to $task_2$ and issue the call $f_2$, which also incurs its own latency. When the result of $f_1$ becomes available, the \textit{Assistant} can resume $task_1$ and continue with the next dependent call. This process continues until all tasks are completed.

\begin{figure*}[!t]
    \centering
    \includegraphics[width=\textwidth]{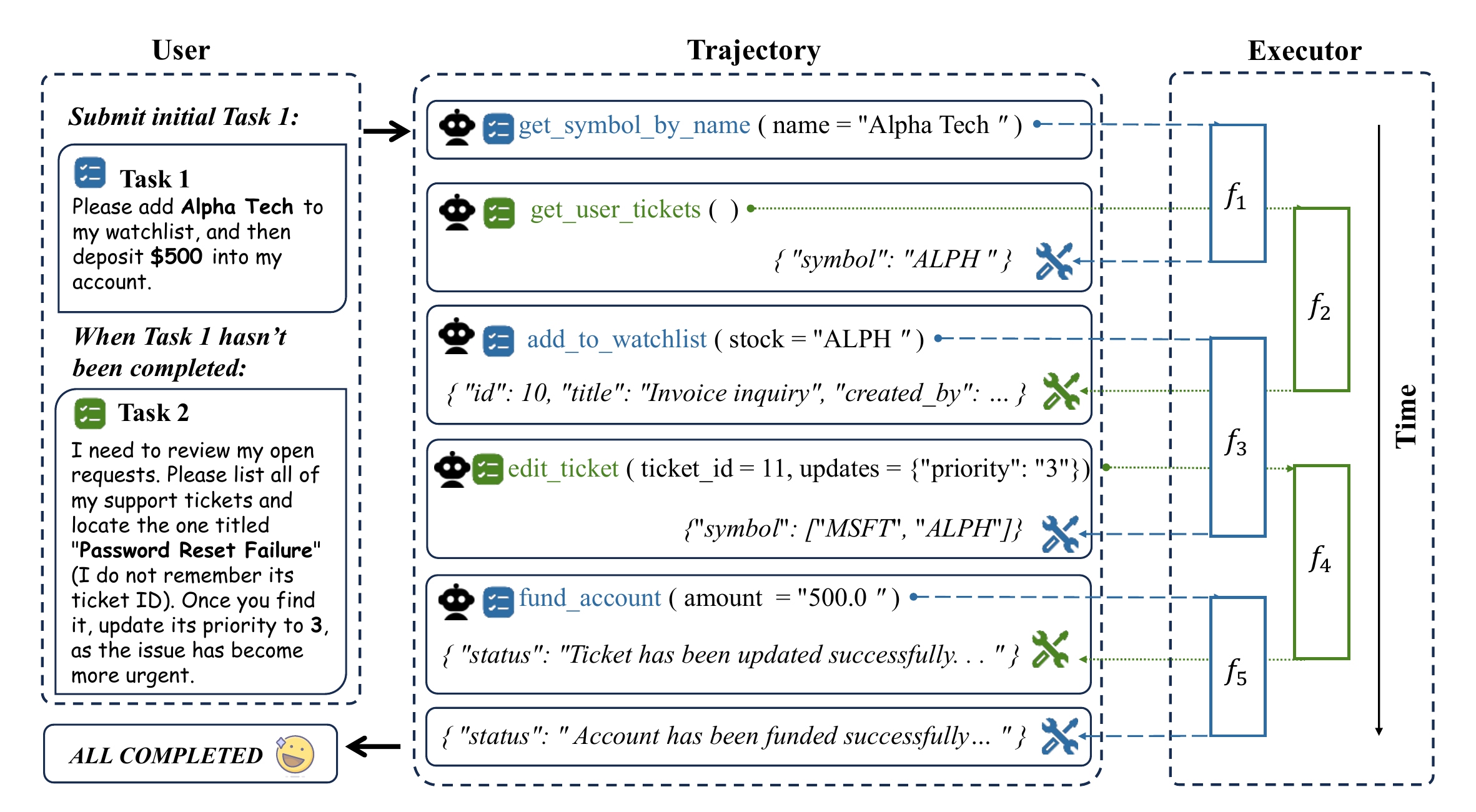}
    \caption{\label{fig:executor}
\textbf{}An example of asynchronous multi-task tool use in \benchname. 
The agent receives two tasks simultaneously and interleaves their dependency-constrained tool-call trajectories under delayed executor feedback. 
When a tool result is pending, the agent can switch to another independent task; when the result becomes available, it resumes the corresponding task. 
The example highlights task interleaving, dependency tracking, and temporal coordination.
}
    \vspace{-1.5em}
\end{figure*}

Figure~\ref{fig:executor} illustrates this interaction process. Rather than evaluating tool use as a purely sequential procedure, {\benchname} evaluates whether an LLM can act as a coordinator that schedules tool calls across multiple pending tasks. A capable agent should not only invoke the correct tools with valid arguments, but also track task states, respect intra-task dependencies, and determine when to switch between tasks under delayed feedback. Consequently, {\benchname} provides a testbed for evaluating temporal coordination and asynchronous task management in tool-using agents.

\subsection{Data Construction}
The construction of {\benchname} requires a high-quality multi-task dataset. To this end, we design a data construction pipeline consisting of four main stages: \textbf{Data Collection} (\S~\ref{subsubsec:data_collection}), \textbf{Coarse Reconstruction} (\S~\ref{subsubsec:reconstruction}), \textbf{Fine-Grained Annotation} (\S~\ref{subsubsec:annotation}), and \textbf{Multi-Task Composition} (\S~\ref{subsubsec:composition}). An overview of the dataset construction process is shown in Figure~\ref{fig:data_construction}.

\subsubsection{Data Collection}
\label{subsubsec:data_collection}

Existing benchmarks have already collected tool APIs derived from real-world scenarios and provide well-developed tool executors, task descriptions, and execution paths. To avoid reinventing the wheel, we leverage these resources as high-quality sources of single-task data. Specifically, we select two representative benchmarks, NESTFUL~\citep{basu2024nestfulbenchmarkevaluatingllms} and BFCLv3~\citep{berkeley-function-calling-leaderboard}.

After automated verification, we categorize and organize the tools and tasks from these benchmarks, ensuring that each task is uniquely associated with a specific tool category. Through this process, we extract a total of 12 tools and 358 tasks, each paired with its corresponding tool-call path, to form the \textbf{Original Dataset}.

\subsubsection{Coarse Reconstruction}
\label{subsubsec:reconstruction}

To ensure reliable evaluation, we first generate ground-truth tool-call trajectories and verify them at both the trajectory and final-environment levels. To reduce manual annotation cost, we then use Gemini 2.5 Pro~\citep{comanici2025gemini} for coarse reconstruction.

Specifically, given the original task description, the multi-step execution trajectory, and the tool set~\citep{hsieh2023tool}, Gemini 2.5 Pro is prompted to reconstruct task descriptions and produce strictly ordered function-call trajectories that align with the reconstructed tasks. We process instances in batches and use carefully designed few-shot prompts to improve the consistency and reliability of the generated data. The detailed prompt is provided in Appendix~\ref{app:analysis}.

Although most reconstructed instances satisfy our requirements, some errors remain, including incorrect function arguments and mismatches between task descriptions and execution orders. We manually inspect and correct these cases to ensure the quality of the final benchmark data.

\subsubsection{Fine-grained Annotation}
\label{subsubsec:annotation}
After refinement with Gemini 2.5 Pro, we obtain a \textbf{Preliminary Dataset}, which may still contain potential issues, including errors that could invalidate an entire tool-call trajectory. To address these quality concerns, we design a fine-grained human annotation pipeline to identify and correct subtle errors and logical inconsistencies introduced during model-based generation.

\textbf{Trajectory Validation.} We first ensure that every function call in a trajectory is valid. To this end, we manually verify the sequential execution results for each task trajectory. Through this process, we identified three recurring error patterns and applied targeted corrections: (1) misinterpretation of the initial task conditions, leading to errors in the first call, such as repeatedly executing cd to enter the current directory in file system tasks; (2) violation of dependency relations, i.e., failing to invoke prerequisite functions, for example skipping a preceding call that is tied to the current one; and (3) misunderstanding of tool functionalities, often manifested as providing function arguments in unsupported or invalid formats.

\textbf{Correction and Disambiguation.} 
Once the validation step confirms that all trajectories are free of execution errors, our focus shifts to aligning tasks with their corresponding trajectories and eliminating ambiguities. First, we verify the consistency between each task and its trajectory, removing task descriptions or partial trajectories that cannot be matched. Second, we strictly enforce the order of function calls within each trajectory, correcting any incorrect sequences. Finally, we replace ambiguous descriptions with precise expressions wherever possible, ensuring that essential details (\eg, location, time, and other key arguments) are explicitly included in the task description.

Following this pipeline, we conduct multiple rounds of verification on 358 tasks until no errors remain, ultimately producing a high-quality \textbf{Single-Task Dataset} comprising 358 validated instances, as summarized in Table \ref{table:data_statics}.

\subsubsection{Multi-task Composition}
\label{subsubsec:composition}

To evaluate agents in realistic multitask scenarios, we consider two factors: task quantity and task type. Task quantity includes dual-task and tri-task settings, while task type includes within-class and cross-class combinations. Combining these factors yields four multitask configurations, which are applied to the single-task dataset.

Since exhaustive combination would produce too many samples, we use weighted random sampling to construct a fixed-size subset. The final \textbf{Multitasking Dataset} contains 712 instances, covering diverse and complex multitasking scenarios.
\subsection{Evaluation}

\begin{table*}[ht]
    \centering
    \caption{\textbf{Main Results of {\benchname}.} Func. and Param. mean matching the model's results with the ground truth to calculate the F1 score. Char. means path matching while Env. means matching multiple execution result environments.
    \label{table:result_main}
    \textbf{Bold} indicates best overall performance, while \underline{underline} denotes the best within the same group. }
    \small
    \begin{tabularx}{\textwidth}{l|*{7}{>{\centering\arraybackslash}X}|>{\centering\arraybackslash}X}
    \toprule
    \multirow{2}{*}{\textbf{Models}}
        & \multicolumn{2}{c}{\textbf{Step Level}} 
        & \multicolumn{3}{c}{\textbf{Sub-Task Level}} 
        & \multicolumn{2}{c|}{\textbf{Task Level}} 
        & \multirow{2}{*}{\textbf{Overall}} \\
    \cmidrule(lr){2-3} \cmidrule(lr){4-6} \cmidrule(lr){7-8}
        & Func. & Param. 
        & Char. & Env. & Acc. 
        & Char. & Env. 
        &  \\ 
    \midrule
    \multicolumn{9}{c}{\textit{Closed-Sourced Models}} \\
    \midrule
    Qwen-Max  & 86.22 & 73.62 & 53.45 & 72.98 & 52.44 & 26.97 & 50.14 & 25.56 \\
    Kimi-K2  & 96.14 & 80.46 & 59.61 & 76.06 & 56.79 & 28.23 & 51.69 & 24.44 \\
    Gemini 2.5 Pro  & 89.08 & 78.27 & 63.16 & 76.43 & 62.05 & 34.13 & 54.35 & 32.44 \\
    GPT-5    & 92.21 & 80.11 & 62.13 & 78.34 & 60.67 & 33.15 & 58.43 & 31.32 \\
    GPT-4o    & 93.92 & 82.26 & 62.90 & 80.04 & 61.41 & 33.71 & 60.53 & 31.74 \\
    GPT-4.1   & \textbf{\underline{96.22}} & \textbf{\underline{84.08}} & \textbf{\underline{67.89}} & \textbf{\underline{82.91}} & \textbf{\underline{67.14}} & \textbf{\underline{39.04}} & \textbf{\underline{64.89}} & \textbf{\underline{38.06}} \\
    \midrule
    \multicolumn{9}{c}{\textit{Open-Sourced Models $<$ 20B}} \\
    \midrule
    LLaMA-3.1-8B-Ins.   & 78.29 & 43.69 & 12.95 & 41.08 & 12.47 & 1.26 & 14.61 & 1.26 \\
    GLM-4-9B-chat       & 74.72 & 59.70 & 29.67 & 55.94 & 28.93 & 7.30 & 25.56 & 6.88 \\
    Qwen2.5-7B-Ins.     & 82.40 & 65.01 & 28.66 & 55.15 & 26.38 & 6.74 & 25.84 & 6.04 \\
    Qwen2.5-14B-Ins.    & 81.32 & 70.22 & \underline{50.21} & 65.29 & 46.28 & \underline{22.61} & 38.20 & \underline{18.82} \\
    Qwen3-8B            & 63.05 & 53.61 & 29.88 & 53.08 & 29.30 & 11.10 & 28.65 & 10.67 \\
    Qwen3-14B           & \underline{85.02} & \underline{72.67} & 49.62 & \underline{69.27} & \underline{47.19} & 20.51 & \underline{44.66} & \underline{18.82} \\
    \midrule
    \multicolumn{9}{c}{\textit{Open-Sourced Models $>$ 20B}} \\
    \midrule
    LLaMA-3.1-70B-Ins.  & 89.60 & 47.10 & 18.47 & 43.26 & 17.83 & 3.37 & 16.43 & 2.81 \\
    LLaMA-3.3-70B-Ins.  & 73.00 & 40.32 & 21.92 & 44.37 & 20.54 & 6.18 & 18.26 & 5.34 \\
    GLM-4-32B    & 60.59 & 51.41 & 35.46 & 56.21 & 33.97 & 16.57 & 29.78 & 15.17 \\
    Qwen3-32B    & 79.95 & 70.37 & 48.14 & 67.46 & 46.71 & 20.22 & 41.43 & 19.10 \\
    Qwen2.5-32B-Ins.    & 94.24 & \underline{81.73} & \underline{61.41} & \underline{74.42} & \underline{56.48} & 30.62 & \underline{49.72} & 24.86 \\
    Qwen3-30B-A3B-Ins. & \underline{94.29} & 80.03 & 54.14 & 72.72 & 53.03 & 22.33 & 47.47 & 21.49 \\
    DeepSeek-V3.1-Terminus    & 86.10 & 75.32 & 59.39 & 72.45 & 56.21 & \underline{33.43} & 49.30 & \underline{28.93} \\
    \bottomrule
    \end{tabularx}
    \vspace{-1.0 em}
\end{table*}

In \textbf{\benchname}, each task is defined as a set of $n$ subtasks 
$\{S_1, S_2, \dots, S_n\}$. Each subtask $S_i$ is represented as a tuple 
$(I_i, Q_i, T_i, E_i)$, where $I_i$ is a unique identifier, $Q_i$ denotes the 
task query, $T_i$ specifies the list of available APIs, and $E_i$ denotes the 
hidden environment state associated with the subtask, which is not directly 
exposed to the assistant. The model's response must explicitly include $I_i$ 
to indicate which subtask is being executed.

For each subtask, we extract its execution trajectory $\mathcal{T}_i$, defined 
as an ordered sequence of tool calls:
\begin{equation}
\label{eq:subtask-trajectory}
\mathcal{T}_i = \langle a_1, a_2, \dots, a_k \rangle,
\end{equation}
where each action $a_j$ is represented as a tuple 
$(\textit{tool}, \textit{args})$. Once all subtasks are completed, we obtain 
the set of trajectories 
$\{\mathcal{T}_1, \mathcal{T}_2, \dots, \mathcal{T}_n\}$, which is then used 
to evaluate whether the model has successfully completed the overall task.

In asynchronous multi-task execution, interactions between the assistant and 
external tools can become highly complex. To provide a comprehensive evaluation 
of the assistant's performance under such conditions, we assess the results at 
three levels: Step Level, Sub-task Level, and Task Level.

\textbf{Step Level.} 
Following the fine-grained evaluation methodology of \citet{patil2023gorilla}, 
we assess the agent's fundamental tool-calling capability, focusing on call 
format, tool selection, and parameter correctness. To quantify these aspects, 
we follow \citet{basu2024nestfulbenchmarkevaluatingllms} and compute F1 scores 
separately for tool accuracy and parameter accuracy.

\textbf{Sub-task Level.} 
At this level, we define accuracy-based metrics to evaluate the agent's 
performance on individual subtasks. For each subtask, we compare the predicted 
trajectory $\mathcal{T}_i^{\mathrm{pred}}$ with the ground-truth trajectory 
$\mathcal{T}_i^{\mathrm{gt}}$ to determine whether the subtask is successfully 
completed, yielding the \textbf{trajectory-completion metric}. In addition, we 
compare the predicted hidden state $E_i^{\mathrm{pred}}$ with the ground-truth 
hidden state $E_i^{\mathrm{gt}}$ to measure environment consistency, yielding 
the \textbf{environment-matching metric}. These two metrics are further 
combined into the \textbf{overall subtask accuracy}, which measures whether a 
subtask is completed both procedurally and environmentally. Detailed 
calculation procedures are provided in Appendix~\ref{appendix:Evaluation_Metrics}.

\textbf{Task Level.} 
At the task level, we evaluate whether the agent successfully completes the 
entire task. The trajectory-completion and environment-consistency metrics at 
this level are counted as correct only when all corresponding subtask-level 
metrics within the task are satisfied. These metrics provide an overall 
assessment of the agent's ability to coordinate and complete multiple subtasks. 
The final task accuracy is defined as the proportion of tasks for which both 
task-level trajectory completion and environment consistency are achieved.

\section{Experiment}

\subsection{Experimental Setup}

We evaluate 19 models on {\benchname}, aiming to provide a comprehensive benchmark for assessing their capability of asynchronous tool calling under multi-task scenarios. Specifically, for closed-source models, we select four prominent models: Qwen-max~\citep{qwen25} created by the Qwen Team, Kimi k2~\citep{team2025kimi} by Kimi Team, Gemini 2.5 Pro ~\citep{comanici2025gemini} developed by Google, alongside GPT-4.1 ~\citep{achiam2023gpt}, GPT-4o ~\citep{hurst2024gpt}, and GPT-5~\citep{gpt5website} by OpenAI.
For open-source LLMs, we evaluate numerous models including LLaMA3.1 ~\citep{llama3modelcard}, LLaMA3.3, Qwen2.5~\citep{qwen2, qwen2.5},  Qwen3 ~\citep{yang2025qwen3}, GLM4 ~\citep{glm2024chatglm}, DeepSeek~\citep{liu2024deepseek}.

\subsection{Results on \benchname}
We conducted a comprehensive empirical evaluation across a wide spectrum of current mainstream models to assess their capabilities. Based on these findings, our analysis is structured around three key questions.

\noindent\textbf{Q1: Which Model is Better in Completing Multiple Tasks Asynchronously?}

As shown in Table \ref{table:result_main}, GPT-4.1 demonstrates the strongest performance in asynchronous capability under asynchronous multitasking evaluation, achieving a score of 38.06. Close behind, the large open-source model DeepSeek-V3.1-Terminus achieves performance highly comparable to that of closed-source models, highlighting its strong competitive capability.

In the step-level evaluation, closed-source models consistently achieve high scores, while open-source models exhibit notable discrepancies. This highlights the differences in the asynchronous capabilities of these models. In the sub-task evaluation, the models' scores are nearly double those of the overload models.

Furthermore, as shown in Appendix \ref{table:result_avg_turns}, the average number of dialogue turns for closed-source models was significantly lower than that of open-source models, which also demonstrates that more powerful models are more efficient in the same environment.

\begin{figure*}[t]
    \centering
    \includegraphics[width=\textwidth]{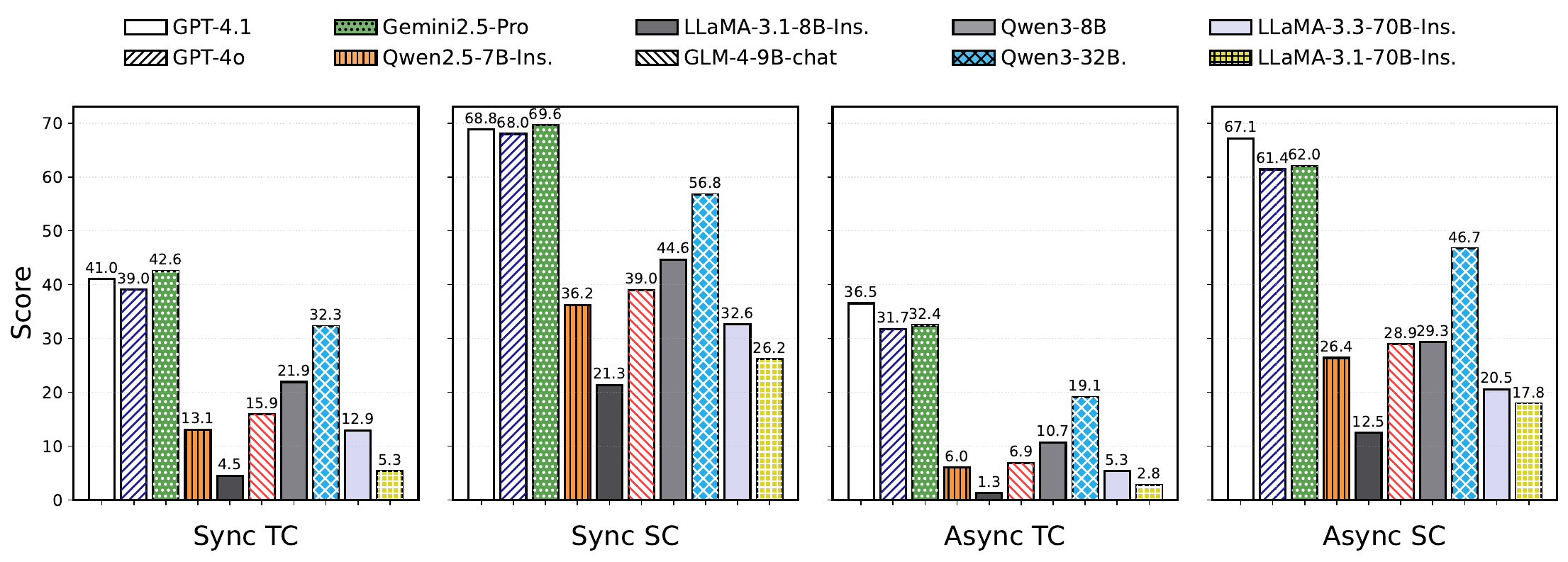}
    \caption{
    Scores of some models. TC is overall Score, and SC is Sub-Task Level Score.
    As the model size reduces, the model score correspondingly declines.
    A higher subtask completion rate does not invariably result in a higher overall score.
    }
    \label{fig:score_SC_TC}
\end{figure*}

\noindent\textbf{Q2: How do Accuracy and Efficiency Trade off in Asynchronous Multi-Task Tool Use?}

Same-task Streak measures the longest consecutive sequence of turns in which the model continues working on the same task, averaged across all samples. A lower value suggests stronger interleaving ability in asynchronous multi-task execution.

\begin{figure}[tb]
    \centering
    \includegraphics[width=\columnwidth]{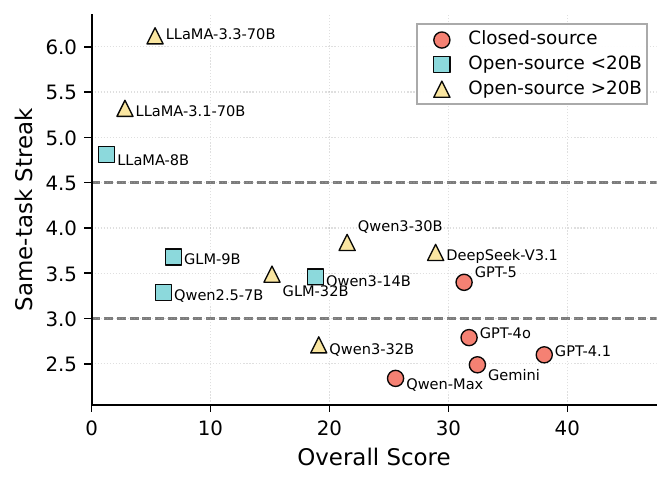}
    \caption{\label{fig:overrall_vs_streak}Trade-off between task accuracy and scheduling efficiency. The x-axis reports Overall Score, while the y-axis reports Same-task Streak. Lower Same-task Streak indicates stronger interleaving ability in asynchronous multi-task execution.}
\end{figure}

Figure \ref{fig:overrall_vs_streak} shows that accuracy and efficiency are not strictly aligned in asynchronous multi-task tool use. The ideal model should appear in the lower-right region, achieving a high Overall score while maintaining a low Same-task Streak, meaning that it can complete tasks accurately and interleave different subtasks efficiently. Closed-source models generally occupy this favorable region: GPT-4.1 achieves the highest Overall score while keeping a relatively low Same-task Streak, indicating that it balances task correctness and asynchronous scheduling most effectively. Gemini 2.5 Pro and GPT-4o also show strong accuracy with compact task switching, suggesting that high-performing models can use idle waiting time to make progress on other tasks rather than repeatedly staying on the same task.

However, Figure \ref{fig:overrall_vs_streak} also reveals a clear trade-off. Some models obtain relatively low Same-task Streak values but still have limited Overall scores, such as smaller open-source models. This suggests that frequent switching alone does not guarantee successful asynchronous execution: models must also correctly track task states, dependencies, and tool outputs. Conversely, models such as DeepSeek-V3.1 achieve competitive Overall scores but with a higher Same-task Streak, indicating stronger task-solving ability but less compact interleaving behavior. Therefore, asynchronous tool-use performance depends on both dimensions: accuracy determines whether the model can complete the required tool trajectories, while efficiency reflects whether it can schedule multiple tasks without wasting waiting time. Overall, Figure 3 suggests that the best models are not simply those that switch most often, but those that switch at the right time while preserving task correctness.

\noindent\textbf{Q3: What Challenges do LLMs Encounter in \benchname?}

In the {\benchname} evaluation, our analysis of tool-call trajectories across various tasks reveals a critical gap in temporal reasoning. Models with lower performance often exhibit a lack of temporal reasoning ability, executing the next function call immediately without waiting for the tool’s response to the current task. This behavior is particularly problematic when dependencies exist between the two calls, often leading to unforeseen errors. In contrast, higher-performing models are able to identify such dependencies, execute tasks sequentially, and leverage idle time to advance other tasks, resulting in a substantial performance gap.

We also observe that agents occasionally fail to complete certain tasks. Specifically, some models tend to execute the most recently presented task first, neglecting earlier tasks. This issue is more common in smaller models and is almost absent in larger 70B-scale models. Moreover, it occurs more frequently in tri-task combinations than in dual-task combinations, which is consistent with intuition. Another notable error is tool misidentification, in which models misjudge the tool use across tasks, for example, invoking a flight-booking tool when they should continue a data-processing task. This type of error typically leads to cascading failures, as the model often struggles to self-correct once the confusion occurs. Common errors, by contrast, include non-compliant instructions, erroneous tool-call trajectories, and parameter errors, all of which directly prevent tasks from being scored successfully. Such errors still constitute a significant proportion of failures in smaller models within {\benchname}, but their frequency decreases substantially as model size increases.

\noindent\textbf{Q4: What Factors Make Multi-Tasking Hard in Asynchronous Tool Calls?}

As shown in Figure \ref{fig:score_SC_TC}, the results show that asynchronous multitasking causes varying degrees of performance degradation across different models compared with regular multitasking, accompanied by a simultaneous decline in the SC index. In terms of the SC metric, Gemini 2.5 Pro exhibits the largest drop among closed-source models, while Qwen3-8B experiences the greatest decrease among open-source models.

The key distinction between asynchronous and regular multitasking is whether the agent can immediately obtain tool responses. In asynchronous settings, each function call incurs a delay, so after invoking a tool, the agent cannot access the information needed for the next step, such as file operation confirmations, query results, or critical state data. High-performing models can handle this uncertainty by shifting their attention to other tasks and making progress via different tool calls while waiting. In contrast, lower-performing models often lack temporal awareness and prematurely advance within the same task, sometimes fabricating parameters based on assumed outcomes from prior calls—an instance of the hallucination behavior commonly observed in large language models.


Another factor influencing difficulty is \textbf{task quantity}. An intuitive hypothesis is that as the number of tasks increases, the overall difficulty also increases. As shown in Figure \ref{fig:heat_map_task_num}, when the number of tasks reaches four, the difficulty of each task increases significantly, and the impact of task quantity becomes more pronounced. Consequently, we do not consider a larger number of tasks in this analysis. Based on {\benchname}, we find that when the number of tasks goes from two to three, the closed-source model Gemini 2.5 Pro decreases the most, from 42.56 to 32.44 while the open-source model Qwen3-8B decreases the most, from 27.91 to 10.67 while the closed-source model GPT-4.1 and the open-source model LLaMA-3.1-70B-Ins decrease less. Full results are listed in Appendix \ref{appendix:appendix_detailed_results}

Other factors include task response latency. As shown in Appendix~\ref{table:resule_time2}, longer latency delays tool execution, resulting in more interactions and increased task difficulty. We additionally report results under random latency settings in Appendix~\ref{table:result_time01} and Appendix~\ref{table:result_time12}; although this introduces additional randomness into the evaluation, it provides a more realistic assessment of model performance.

\begin{figure}[tb]
    \centering
    \includegraphics[width=\columnwidth]{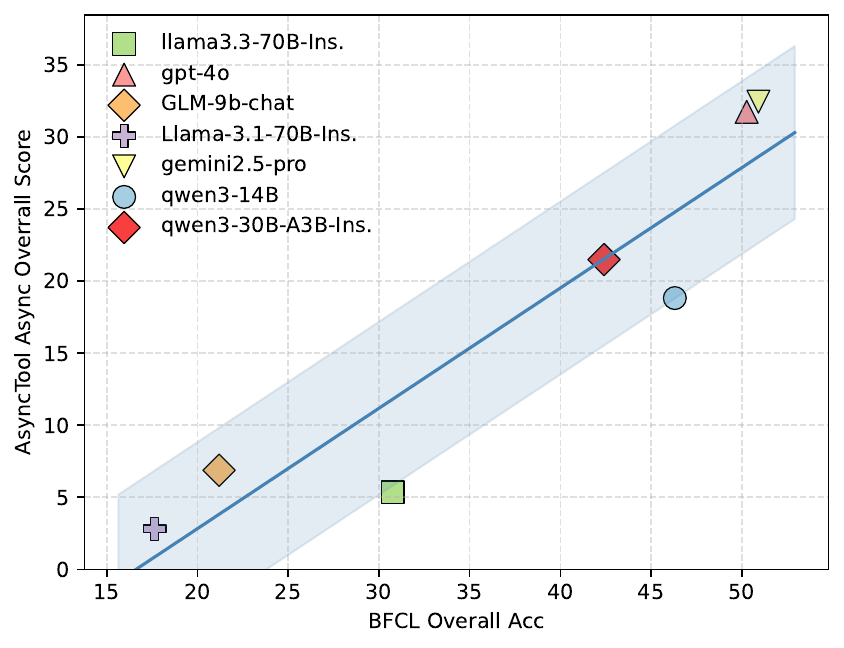}
    \caption{Comparison between BFCL Overall Accuracy and \benchname\ Overall Scores across several models.}
    \label{fig:bench_compare}
\end{figure}

\section{Conclusion}

We present \benchname, a benchmark for evaluating asynchronous tool calling in interactive multi-task environments with delayed tool feedback. {\benchname} differs from conventional tool-use evaluations by requiring agents to coordinate multiple tasks, utilize idle waiting periods, and respect intra-task dependencies when tool results are not immediately available. Based on validated single-task tool-use trajectories, we construct a diverse asynchronous multitasking dataset and evaluate model behavior at the step, sub-task, and task levels, together with efficiency-oriented metrics for task interleaving. Extensive experiments show that asynchronous execution remains challenging for current LLM-based agents, causing clear performance degradation under delayed feedback. Our analysis shows that effective asynchronous tool use requires correct tool calls, task-state maintenance, dependency tracking, and timely task switching. Strong agents should switch tasks at appropriate moments rather than frequently. We hope \textsc{AsyncTool} can support future research on temporally aware, efficient, and reliable tool-using agents.

\section*{Limitations}
In \benchname, we introduce response delays into tool calls for the first time, enabling a fine‑grained evaluation of tool‑using agents’ performance in asynchronously executing multiple tasks. Nonetheless, two notable limitations remain. First, our dataset is reconstructed from BFCLv3 and NESTFUL. Although tasks are carefully annotated at a fine‑grained level, the task scenarios and the range of available tools are inherently constrained by these sources. Second, the data reconstruction process relies on the high‑performance Gemini‑2.5‑pro, making dataset construction costly. This dependency not only restricts further expansion of the dataset but also introduces inevitable generation errors that require additional human verification, thereby increasing annotation costs. These limitations present challenges to the diversified scaling of asynchronous multitasking evaluation.

\section{Acknowledgement}

This work was supported by the Anhui Provincial Natural Science Foundation under Grant 2108085UD12. We acknowledge the support of GPU cluster built by MCC Lab of Information Science and Technology Institution, USTC. The AIdriven experiments, simulations and model training were performed on the robotic AI-Scientist platform of the Chinese Academy of Sciences.
\bibliography{asynctool}

\clearpage

\appendix
\section{Related Work}
\textbf{Multi-Step Tool Call.}
As LLM-based agents continue to improve in their ability to leverage external tools, related benchmarks have higher complexity and richer dependencies. Some tasks now require multiple tool calls, executed in the correct order, to be completed, which presents significant challenges for the models in terms of comprehension and planning \citep{chen2025tool, yin2025toolvqa, xie2024travelplanner, zheng2024natural}. Currently, two primary approaches are used to evaluate this capability. The first approach asks the model to provide, in a single step, the full sequence of tool calls needed to solve the task, focusing on the model's ability to plan tool usage \citep{huang-etal-2024-planning-creation, yu2025multimissiontoolbenchassessing, zhang-etal-2024-reverse, chen-etal-2024-eval, 10.1007/978-981-97-9434-8_29}. The second approach allows the model to invoke tools step by step, receiving responses at each stage. It emphasizes evaluating the multi-step tool calling process until task completion \citep{berkeley-function-calling-leaderboard, basu2024nestfulbenchmarkevaluatingllms, zhong2025complexfuncbench}.

The order of tool calls is often a critical aspect to evaluate. \citep{berkeley-function-calling-leaderboard} constrains the tool call sequence using traditional rule-based restrictions, while NESTFUL \citep{basu2024nestfulbenchmarkevaluatingllms} enforces a fixed call order through parameter dependencies. 
A key limitation of existing benchmarks is their failure to account for tool invocation delays frequently encountered in real-world applications. {\benchname} addresses this gap, enabling systematic evaluation of a model’s capacity for asynchronous tool calling.

\textbf{Asynchronous Multitasking Scenarios.}
The ability to perform multiple tasks asynchronously is an inevitable trend in the development of large LLMs. \citep{chu2023timebench,tan2023towards,wang2023tram, ginart2024asynchronous, hettige2025modular, chai2025survey, wu2025recipe2plan} introduce time changes in model evaluation and provide related evaluation tasks.\citep{zhang2024timearena} introduces the concept of time in multitasking, aiming to assess the efficiency with which models manage multiple tasks concurrently. \citep{gonzalez2025robotouille} evaluates models' asynchronous planning, focusing on failure modes and challenges in integrating long-term information. While important for understanding LLMs in asynchronous tasks, these studies are limited to simulations and don't assess performance on tasks requiring real-world tool calls.
 Moreover, \citep{gim2024asynchronous} enhances the operational efficiency of LLMs by enabling them to generate and execute function calls concurrently, representing a novel approach. Unlike prior benchmarks, {\benchname} is the first to systematically evaluate models’ capabilities for asynchronous tool calling in realistic multi-task scenarios, explicitly incorporating tool response latency into evaluation environment.

\section{Evaluation}
\numberwithin{equation}{section}
\setcounter{equation}{0}
\subsection{Metrics}\label{appendix:Evaluation_Metrics}

\textbf{Character matching.} We extract the function name and parameters and match them against the golden truth.
We also consider the possibility that the model might execute some incorrect function calls due to the use of subset validation.
Specifically, given a tool list $T$ and a query $q$, the model generates a series
\begin{equation}
\label{eq:predicted-trajectory}
\mathcal{T}^{\text{pred}} = \langle a_1^{\text{pred}}, a_2^{\text{pred}}, \dots, a_{n}^{\text{pred}}\rangle,
\end{equation}
where $a_i^{\text{pred}}$ is the predicted LLM action at turn $i$.
The golden truth is
\begin{equation}
\label{eq:ground-truth-trajectory}
\mathcal{T}^{\text{gt}} = \langle a_1^{\text{gt}}, a_2^{\text{gt}}, \dots, a_{n}^{\text{gt}}\rangle.
\end{equation}
We verify whether $A^{\text{gt}} \subseteq A^{\text{pred}}$.

\textbf{F1 function \& Parameters.} Referring to the implementation of NESTFUL(\cite{basu2024nestfulbenchmarkevaluatingllms}), we use the above metrics to evaluate the model's responses. In our experiments, we find that frequent task switching by the model interferes with its ability to recognize different tasks accurately. Specifically, the model often invokes incorrect tools for the tasks, which leads to a decline in performance metrics.

\textbf{Trajectory completion.} After we extract the function, we will throw $\mathcal{T}^{gt}$ and $\mathcal{T}^{pred}$ to the executor for execution. The executor will return the corresponding results $R^{gt}$ and $R^{pred}$, and we will compare the results using the subset judgment method the same as above.

\textbf{Environment matching.} Referring to the implementation of BFCL(\cite{berkeley-function-calling-leaderboard}), we use the executor class instance for comparison. In our data, some tool calls induce alterations to the environment, while the majority of tool calls do not result in any environmental changes, such as those associated with queries. Nonetheless, we incorporate this metric into the evaluation to ensure comprehensiveness.

\subsection{Compare {\benchname} with other benchmarks}
Table~\ref{table:compare_benchmarks} shows how {\benchname} compares against
existing tool-use and asynchronous benchmarks.

\begin{table*}[t!]
\centering
\footnotesize
\caption{\textbf{Comparison between {\benchname} and existing benchmarks.}}  
\label{table:compare_benchmarks}
\begin{tabular}{cccccc}
\toprule
\textbf{Benchmark} & \textbf{Async Executor} & \textbf{Function Call} & \textbf{Multi Task} & \textbf{Multi Step} & \textbf{Cross Scenario} \\
\midrule
$\tau$-bench~\citep{yao2024taubenchbenchmarktoolagentuserinteraction} & \xmark & \cmark & \xmark & \xmark & \xmark  \\
BFCL v3~\citep{berkeley-function-calling-leaderboard} & \xmark & \cmark & \xmark & \cmark & \xmark \\
NestFul~\citep{basu2024nestfulbenchmarkevaluatingllms} & \xmark & \cmark & \xmark & \cmark & \xmark \\
TimeArena~\citep{zhang2024timearena} & \cmark & \xmark & \cmark & \cmark & \xmark  \\
C\^{}3-Bench~\citep{yu2025c} & \xmark & \cmark & \cmark & \cmark & \xmark  \\
Robotouille~\citep{gonzalez2025robotouille} & \cmark & \xmark & \xmark & \cmark & \xmark \\
\midrule
\textbf{\benchname\ (Ours)} & \cmark & \cmark & \cmark & \cmark & \cmark \\
\bottomrule
\end{tabular}
\end{table*}

\begin{figure}[tb]
    \centering
    \includegraphics[width=\columnwidth]{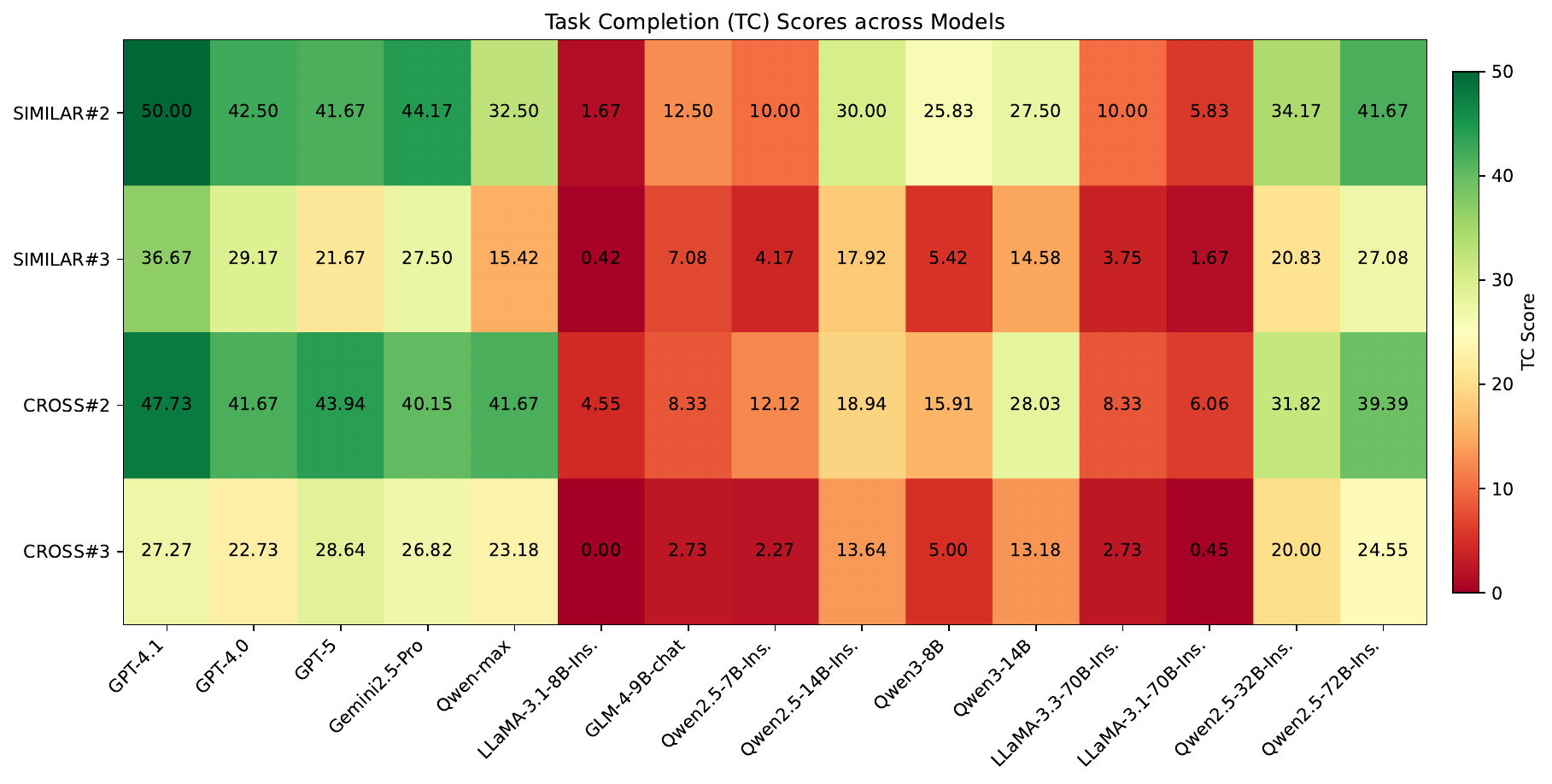}
    \caption{\label{fig:heat_map_task_num}The performance of different models on specific task classifications. Red indicates poorer performance, and green indicates better performance.}
\end{figure}
\begin{table}[ht]
\centering
\centering
  \caption{\textbf{Abbreviation explanation of data categories.}}
  \label{table:data_type}
  \begin{tabular}{lcc}
    \toprule
    \textbf{Full Name} & \textbf{Abbreviation} \\
    \midrule
    Data Management & DM \\
    Filesystem & FS \\
    Data generation & DG \\
    MessageAPI & MA \\
    Number operations & NO \\
    SocialConnect & SC \\
    String Manipulation & SM \\
    TicketPurchase & TP \\
    TradingBot & TB \\
    TravelPlanning & TP* \\
    DataFormat & DF \\
    Machine Operation & MO \\
    \bottomrule
  \end{tabular}
\end{table}

\begin{table}[!t]
    \centering
    \small
    \caption{\textbf{Task Counts and Average Trajectory Lengths.}}
    \label{table:data_statics}
    \resizebox{0.95\linewidth}{!}{
    \begin{tabular}{lcccccccccccc}
    \toprule
    \textbf{Type} & \textbf{DM} & \textbf{FS} & \textbf{DG} & \textbf{MA} & \textbf{NO} & \textbf{SC} & \textbf{SM} & \textbf{TP} & \textbf{TB} & \textbf{TP*} & \textbf{DF} & \textbf{MO} \\
    \midrule
    \#Tasks & 59 & 53 & 29 & 23 & 24 & 18 & 34 & 21 & 15 & 20 & 44 & 18 \\
    Avg. len. & 2.14 & 3.58 & 2.14 & 3.48 & 2.12 & 4.11 & 2.09 & 2.81 & 3.27 & 3.10 & 2.09 & 3.00 \\
    \bottomrule
    \end{tabular}
    }
\end{table}

\subsection{Model Version}
The version for GPT-4o is \texttt{gpt-4o-2024-11-20}, for GPT-4.1 is \texttt{gpt-4.1-2025-04-14}, and for Gemini 2.5 Pro is \texttt{gemini-2.5-pro-preview-05-06}. The version for GLM-4-32B is \texttt{GLM-4-32B-0414}. The version for Qwen3-30B-A3B-Ins. is \texttt{Qwen3-30B-A3B-Instruct-2507}.

\section{Data}

\subsection{Human Annotation Instructions}

We used human annotators to verify and refine the single-task tool-use trajectories during dataset construction. The annotators were instructed to check whether each task description, tool-call trajectory, and execution result were mutually consistent. The full instructions given to annotators are as follows:

\paragraph{Annotation Goal.}
Given a task description, a list of available tools, and a candidate multi-step tool-call trajectory, please determine whether the trajectory correctly solves the task. The goal is to ensure that each trajectory is executable, deterministic, and consistent with the task description.

\paragraph{Annotation Procedure.}
Annotators were asked to follow these steps:

\begin{enumerate}
    \item Read the task description carefully and identify the user's intended goal.
    \item Inspect the ordered tool-call trajectory and check whether each function call is valid.
    \item Verify that the function name and arguments are supported by the corresponding tool.
    \item Check whether the trajectory respects all dependency relations between tool calls. A later call should not use information that has not been obtained from a previous tool response.
    \item Execute or inspect the trajectory results and confirm that the final environment state satisfies the task requirement.
    \item If an error is found, mark the error type and provide a corrected version when the correction is unambiguous.
    \item If the task description is ambiguous or inconsistent with the trajectory, rewrite the description to make key information such as entities, time, location, or required arguments explicit.
    \item Remove instances that cannot be reliably corrected or whose task description cannot be aligned with the execution trajectory.
\end{enumerate}

\paragraph{Common Error Types.}
Annotators were asked to pay special attention to the following errors:
\begin{itemize}
    \item incorrect interpretation of the initial task condition;
    \item missing prerequisite tool calls;
    \item invalid function names or unsupported argument formats;
    \item incorrect ordering of dependent function calls;
    \item mismatch between the task description and the execution trajectory;
    \item ambiguous task descriptions that may lead to multiple valid trajectories.
\end{itemize}

\paragraph{Risk and Privacy Notice.}
The annotation task only involved checking synthetic benchmark task descriptions and tool-call trajectories. Annotators were not asked to provide personal information, opinions, or sensitive data. The annotation process did not involve interaction with real users or real private user data. Therefore, the risk to annotators was minimal.

\paragraph{Quality Control.}
Each instance was checked for trajectory validity, task-trajectory consistency, and final environment correctness. The annotation process was repeated until no execution errors or obvious inconsistencies remained in the validated single-task dataset.

\subsection{Data details}\label{appendix:data_info}
As shown in Table \ref{table:data_type}, the content inside introduces the meaning of our categories. Besides, Table \ref{table:data_info} shows the data distribution of {\benchname}.

\subsection{Data length}
\begin{figure*}[t]
    \centering
    \subfloat[GPT-4.1]{
        \includegraphics[width=0.32\textwidth]{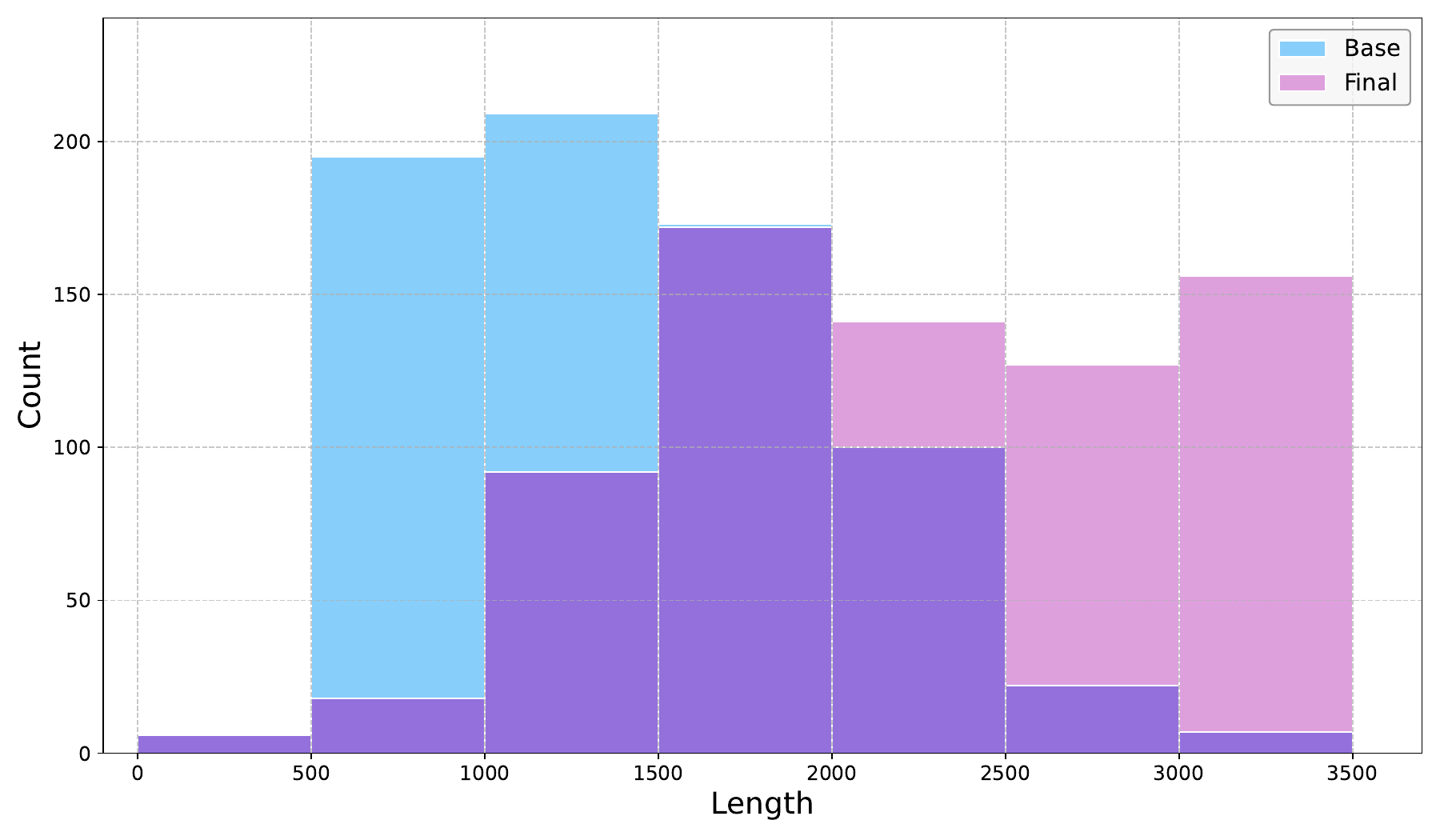}
    }
    \hfill
    \subfloat[GPT-4o]{
        \includegraphics[width=0.32\textwidth]{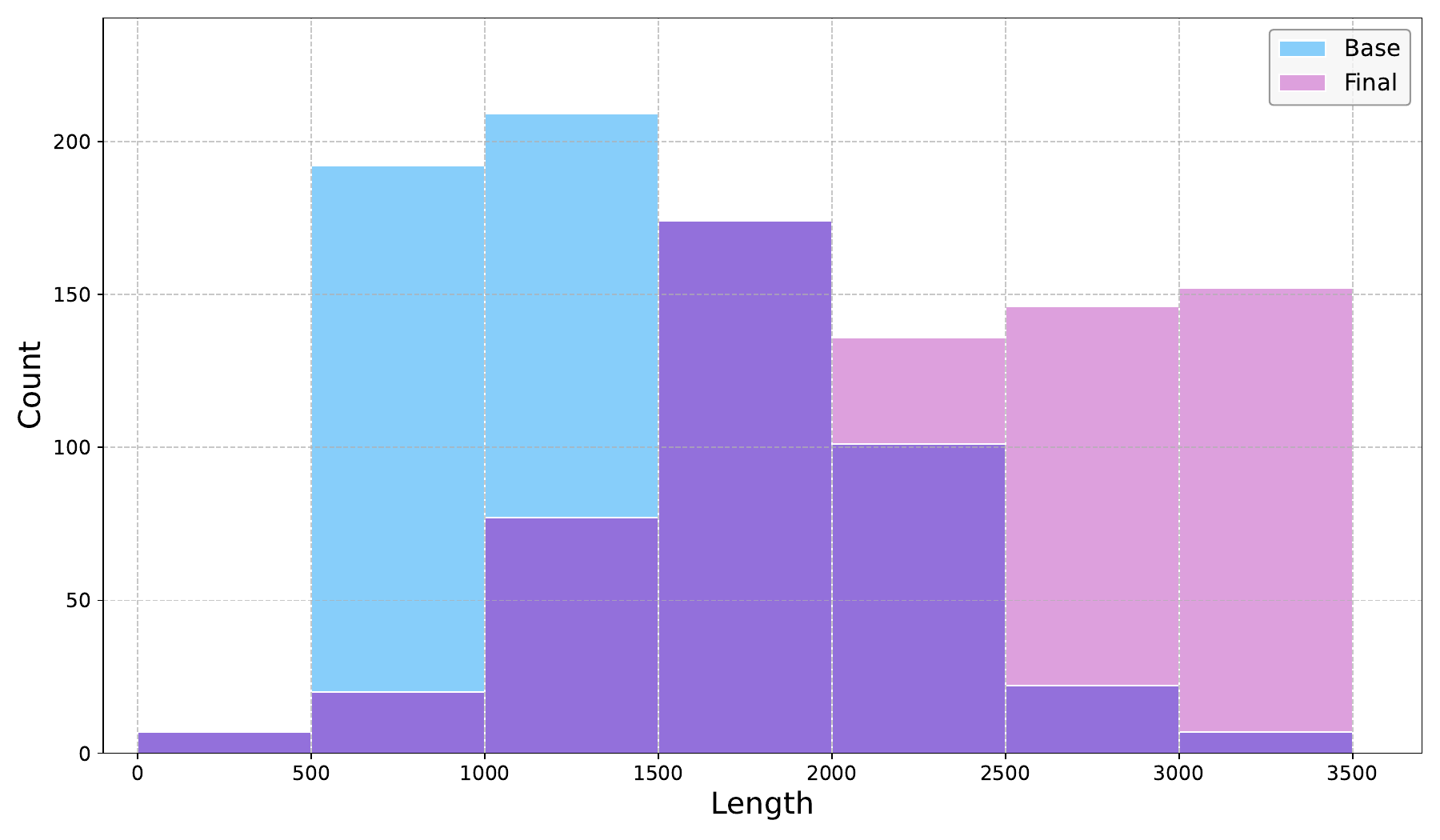}
    }
    \hfill
    \subfloat[Gemini 2.5 Pro]{
        \includegraphics[width=0.32\textwidth]{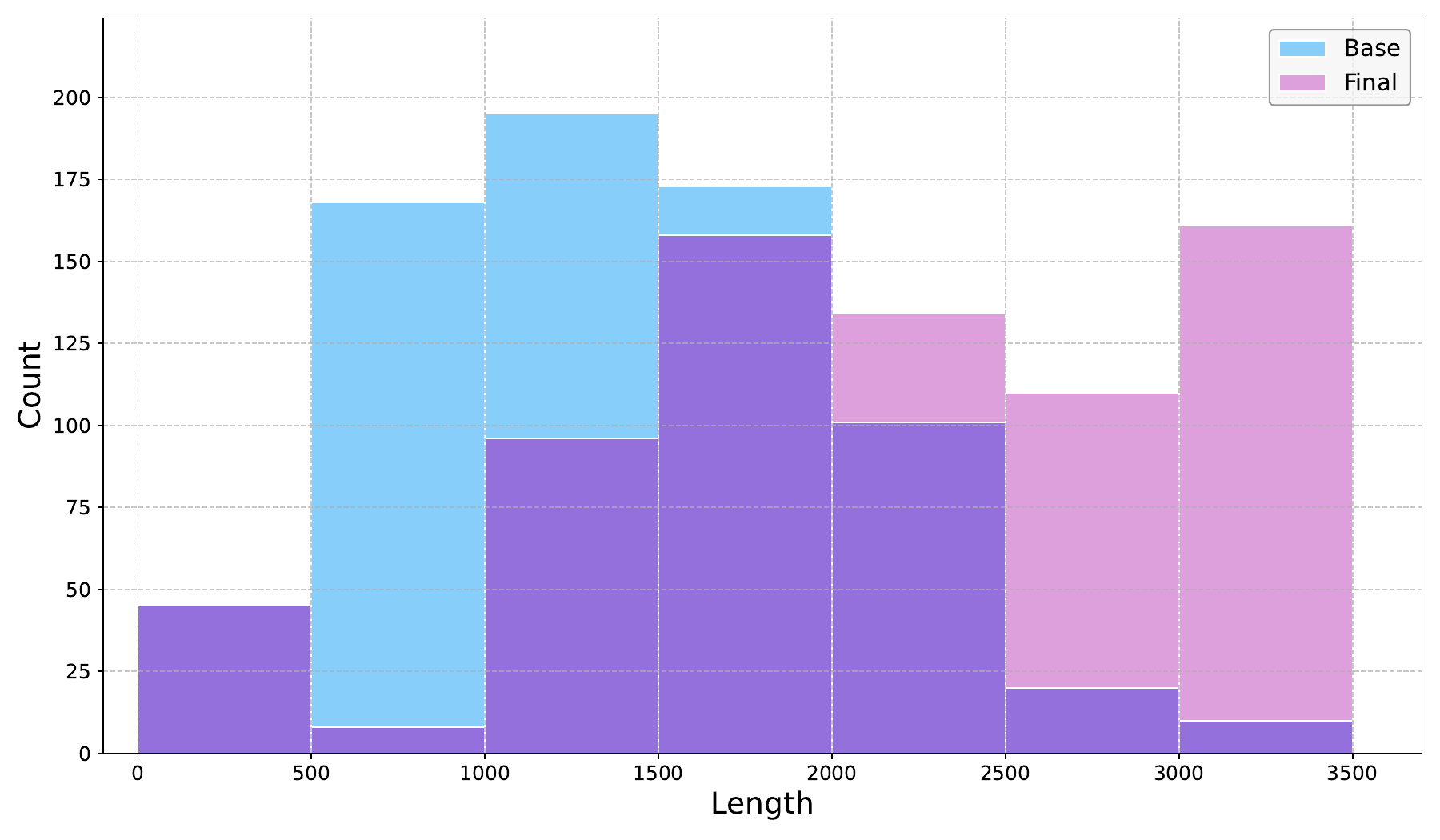}
    }
    \caption{Length distribution for base and final conversation, measured by the number of tokens.}
    \label{fig:token_distribution}
\end{figure*}

We counted the input and output lengths of each task to ensure that they were within a reasonable range. These data are shown in Figure \ref{fig:token_distribution}.

\subsection{Supplementary data}
For the four-task test, considering the large number of combination types and tasks, we randomly screened the data after each combination and finally selected 300 SIMILAR and 300 CROSS four-task data for comparison. The results are only used for ablation experiments.

\begin{table}[ht]
\centering
\centering
  \caption{\textbf{Data composition of ASYNCTOOL.} \# denotes number of, and \% denotes proportion of. SIMILAR means similar task combinations and CROSS means cross task combinations.}
  \label{table:data_info}
  \begin{tabular}{lccc}
    \toprule
    \textbf{Method} & \textbf{\#Tasks} & \textbf{\#Samples} & \textbf{\%Samples} \\
    \midrule
    SIMILAR   & 2  & 120    & 16.85\%   \\
    CROSS   &  2  & 132    & 18.54\%    \\
    SIMILAR   &  3  & 240    & 33.71\%   \\
    CROSS   &  3  & 220    & 30.90\%    \\
    \bottomrule
  \end{tabular}
\label{tab:table1}
\end{table}

\section{Results}\label{appendix:appendix_results}
In this section, we present the comprehensive results, categorized by different types, along with the test outcomes for varying numbers of tasks on the open-source model.
\subsection{Detailed results}\label{appendix:appendix_detailed_results}
As shown in Figure ~\ref{table:result_detailed}, we compile and present the detailed results by task type.
\begin{table*}[ht]
    \centering
    \caption{\textbf{Detailed Results of {\benchname}.} 
    \label{table:result_detailed}We approach the analysis from the perspective of dataset classification to calculate the results. We find that the scores across different categories exhibit significant disparities.
    \textbf{Bold} indicates best overall performance, while \underline{underline} denotes the best within the same group. }
    \small
    \begin{tabularx}{\textwidth}{l|*{4}{>{\centering\arraybackslash}X}|>{\centering\arraybackslash}X}
    \toprule
    \textbf{Models} & \textbf{SIMILAR\#2} & \textbf{SIMILAR\#3} & \textbf{CROSS\#2} & \textbf{CROSS\#3} & \textbf{Overall} \\
    \midrule
    \multicolumn{6}{c}{\textit{\textbf{Closed-Sourced Models}}} \\
    \midrule
    Qwen-max & 32.50 & 15.42 & 41.67 & 23.18 & 25.56 \\
    Kimi-K2 & 33.33 & 20.83 & 30.30 & 20.00 & 24.44 \\
    Gemini 2.5 Pro  & 44.17 & 27.5 & 40.15 & 26.82 & 32.44 \\
    GPT-5        & 41.67 & 21.67 & 43.94 & \textbf{\underline{28.64}}& 31.32 \\
    GPT-4o        & 42.50 & 29.17 & 41.67 & 22.73 & 31.74 \\
    GPT-4.1   & \textbf{\underline{50.00}} & \textbf{\underline{36.67}} & \textbf{\underline{47.73}} & 27.27 & \textbf{\underline{38.06}} \\
    \midrule
    \multicolumn{6}{c}{\textit{\textbf{Open-Sourced Models $<$ 20B}}} \\
    \midrule
    LLaMA-3.1-8B-Ins.   & 1.67 & 0.42 & 4.55 & 0 & 1.26 \\
    GLM-4-9B-chat       & 12.50 & 7.08 & 8.33 & 2.73 & 6.88 \\
    Qwen2.5-7B-Ins.     & 10.00 & 4.17 & 12.12 & 2.27 & 6.04 \\
    Qwen2.5-14B-Ins.    & \underline{30.00} & \underline{17.92} & 18.94 & \underline{13.64} & \underline{18.82} \\
    Qwen3-8B       & 25.83 & 5.42 & 15.91 & 5.00 & 10.67 \\
    Qwen3-14B    & 27.50 & 14.58 & \underline{28.03} & 13.18 & \underline{18.82} \\
    \midrule
    \multicolumn{6}{c}{\textit{\textbf{Open-Sourced Models $>$20B}}} \\
    \midrule
    LLaMA-3.1-70B-Ins.& 5.83 & 1.67 & 6.06 & 0.45 & 2.81 \\
    LLaMA-3.3-70B-Ins.    & 10.00 & 3.75 & 8.33 & 2.73 & 5.34 \\
    GLM-4-32B    & 21.67 & 13.75 & 21.97 & 9.09 & 15.17 \\
    Qwen3-32B    & 31.67 & 13.33 & 26.52 & 14.09 & 19.10 \\
    Qwen2.5-32B-Ins.    & \underline{34.17} & 20.83 & 31.82 & 20.00 & 24.86 \\
    Qwen3-30B-A3B-Ins.    & 30.00 & 15.42 & 32.58 & 16.82 & 21.49 \\
    DeepSeek-V3.1-Terminus    & 32.50 & \underline{24.58} & \underline{35.61} & \underline{27.73} & \underline{28.93} \\
    \bottomrule
    \end{tabularx}
\end{table*}

\subsection{Other results}
As shown in Figure ~\ref{table:compare_tasknum}, we conduct a preliminary experiment on the open-source model to demonstrate the impact of the number of tasks on the score.
\begin{table*}[ht]
    \centering
    \caption{\textbf{Experimental results with different numbers of tasks on SYNC setting.} \\ In \#number,number means the number of tasks.  
    \# 3 is compared with \# 2 and \# 4 is compared with \# 3. The data source of \# 4 is shown in Appendix \ref{appendix:data_info}}
    \label{table:compare_tasknum}
    \scriptsize
    \begin{tabularx}{\textwidth}{l|*{3}{>{\centering\arraybackslash}X}|*{4}{>{\centering\arraybackslash}X}|*{4}{>{\centering\arraybackslash}X}}
    \toprule
    \multirow{2}{*}{\textbf{Models}} & 
    \multicolumn{3}{c|}{\textbf{\# 2}} & 
    \multicolumn{4}{c|}{\textbf{\# 3}} & 
    \multicolumn{4}{c}{\textbf{\# 4}} \\
    \cmidrule(lr){2-4} \cmidrule(lr){5-8} \cmidrule(lr){9-12}
    & S\#2 & C\#2 & Overall 
    & S\#3 & C\#3 & Overall & Drop
    & S\#4 & C\#4 & Overall & Drop \\
    \midrule
    Qwen2.5-7B-Ins. & 20.00 & 14.39 & 17.06 & 11.67 & 10.00 & 10.87 & 36.28\% & 8.51 & 3.44 & 6.33 & 41.77\% \\
    GLM-4-9b-chat & 20.83 & 21.97 & 21.43 & 14.58 & 10.91 & 12.83 & 40.13\% & 12.73 & 3.65 & 8.17 & 36.32\% \\
    \midrule
    Qwen2.5-72B-Ins. & 53.33 & 47.73 & 50.79 & 35.42 & 31.82 & 33.70 & 33.65\% & 20.33 & 23.67 & 22.00 & 34.72\% \\
    LLaMA3.3-70B-Ins. & 25.00 & 16.67 & 20.63 & 10.42 & 6.82 & 8.70 & 57.83\% & 3.67 & 3.67 & 3.67 & 57.81\% \\

    \bottomrule
    \end{tabularx}
\end{table*}

\subsection{Analysis of error cases}
We provide numerous examples of common model errors in the end of appendix. A frequent issue, likely due to insufficient prior training on multiple tasks, is the model forgetting about ongoing tasks. This results in a high rate of incompletion failures, despite the model's high accuracy in completing subtasks. In open-source models, frequent task switching presents an even greater challenge to the model's memory and localization abilities, causing confusion in function calls, which is a primary source of errors.

\subsection{Analysis of model performance}
We must acknowledge that even the most advanced models perform suboptimally on our benchmark. However, we need to analyze the reasons behind this from two perspectives. First, we have conducted ablation experiments on mainstream large models, revealing that as the number of tasks increases, the difficulty grows nonlinearly. To encourage future models to achieve greater capabilities, we designed the benchmark such that tasks involving three or more objectives account for over 60

\subsection{Ablation results}
\begin{table*}[ht]
    \centering
    \caption{\textbf{Supplementary Results of {\benchname}. All experiments are conducted with a delay of two turns.} 
    \label{table:resule_time2}
    \textbf{Bold} indicates best overall performance, while \underline{underline} denotes the best within the same group. }
    \small
    \begin{tabularx}{\textwidth}{l|*{7}{>{\centering\arraybackslash}X}|>{\centering\arraybackslash}X}
    \toprule
    \multirow{2}{*}{\textbf{Models}}
        & \multicolumn{2}{c}{\textbf{Step Level}} 
        & \multicolumn{3}{c}{\textbf{Sub-Task Level}} 
        & \multicolumn{2}{c|}{\textbf{Task Level}} 
        & \multirow{2}{*}{\textbf{Overall}} \\
    \cmidrule(lr){2-3} \cmidrule(lr){4-6} \cmidrule(lr){7-8}
        & Func. & Param. 
        & Char. & Env. & Acc. 
        & Char. & Env. 
        &  \\ 
    \midrule
    \multicolumn{9}{c}{\textit{Closed-Sourced Models}} \\
    \midrule
    Qwen-Max  & \textbf{\underline{95.81}} & \textbf{\underline{82.70}} & \textbf{\underline{64.7}} & \textbf{\underline{77.18}} & \textbf{\underline{60.18}} & \textbf{\underline{35.11}} & 53.51 & \textbf{\underline{30.76}} \\
    Gemini 2.5 Pro  & 91.94 & 80.05 & 57.48 & 75.90 & 56.48 & 26.97 & 53.65 & 25.70 \\
    GPT-4o    & 91.41 & 79.79 & 57.70 & 75.42 & 56.79 & 27.67 & 53.37 & 26.54 \\
    GPT-4.1   &     91.32 & 79.97 & 57.96 & 75.90 & 56.90 & 27.67 & \textbf{\underline{54.49}} & 26.40 \\
    \midrule
    \multicolumn{9}{c}{\textit{Open-Sourced Models $<$ 20B}} \\
    \midrule
    LLaMA-3.1-8B-Ins.   & 83.16 & 44.42 & 13.38 & 40.76 & 12.42 & 1.54 & 14.75 & 1.40 \\
    Qwen2.5-7B-Ins.     & \underline{87.67} & 66.97 & 32.27 & 58.28 & 29.99 & 6.88 & 26.40 & 5.48 \\
    Qwen2.5-14B-Ins.    & 86.75 & \underline{72.18} & \underline{50.21} & \underline{63.59} & \underline{45.06} & \underline{19.94} & \underline{35.11} & \underline{15.17} \\
    Qwen3-8B            & 59.06 & 49.4 & 24.52 & 48.99 & 23.99 & 6.6 & 21.77 & 6.6 \\
    \midrule
    \multicolumn{9}{c}{\textit{Open-Sourced Models $>$ 20B}} \\
    \midrule
    LLaMA-3.1-70B-Ins.  & 90.73 & 45.45 & 17.04 & 40.02 & 16.14 & 1.83 & 12.36 & 1.4 \\
    Qwen3-32B    & 70.67 & 60.99 & 34.34 & 56.95 & 33.7 & 11.52 & 30.76 & 10.96 \\
    Qwen3-30B-A3B-Ins. & \underline{95.57} & \underline{79.72} & \underline{55.41} & \underline{71.71} & \underline{52.81} & \underline{23.17} & \underline{44.66} & \underline{20.65} \\
    DeepSeek-V3.1-Terminus    & 81.69 & 70.52 & 43.84 & 65.02 & 42.94 & 15.73 & 36.94 & 14.61 \\
    \bottomrule
    \end{tabularx}
    \vspace{-1.0 em}
\end{table*}
\begin{table*}[ht]\centering
    \caption{\textbf{Results of{\benchname}. All experiments are conducted with a delay of zero to one turn randomized.}
    \label{table:result_time01}
    \textbf{Bold} indicates best overall performance, while \underline{underline} denotes the best within the same group. }
    \small
    \begin{tabularx}{\textwidth}{l|*{7}{>{\centering\arraybackslash}X}|>{\centering\arraybackslash}X}
    \toprule
    \multirow{2}{*}{\textbf{Models}}
        & \multicolumn{2}{c}{\textbf{Step Level}} 
        & \multicolumn{3}{c}{\textbf{Sub-Task Level}} 
        & \multicolumn{2}{c|}{\textbf{Task Level}} 
        & \multirow{2}{*}{\textbf{Overall}} \\
    \cmidrule(lr){2-3} \cmidrule(lr){4-6} \cmidrule(lr){7-8}
        & Func. & Param. 
        & Char. & Env. & Acc. 
        & Char. & Env. 
        &  \\ 
    \midrule
    \multicolumn{9}{c}{\textit{Closed-Sourced Models}} \\
    \midrule
    Qwen-Max  & \underline{\textbf{96.65}} & \underline{\textbf{84.21}} & \underline{\textbf{66.30}} & 80.57 & 64.65 & \underline{\textbf{36.80}} & 58.43 & 34.41  \\
    Gemini 2.5Pro& 95.57 & 83.85 & 65.45 & 81.69 & 64.92 & 35.81 & 62.22 & 35.25  \\
    GPT-4o& 96.00 & 84.03 & 66.08 & 81.74 & \underline{\textbf{65.61}} & 35.96 & 62.08 & \underline{\textbf{35.53}} \\
    GPT-4.1& 95.45 & 83.58 & 65.76 & \underline{\textbf{81.90}} & 65.18 & 35.67 & \underline{\textbf{62.92}} & 34.55  \\
    \midrule
    \multicolumn{9}{c}{\textit{Open-Sourced Models $<$ 20B}} \\
    \midrule
    LLaMA-3.1-8B-Ins.   & 83.93 & 47.44 & 16.24 & 43.84 & 15.13 & 1.97 & 16.71 & 1.69  \\
    Qwen2.5-7B-Ins.     & 83.73 & 67.22 & 32.11 & 59.02 & 31.00 & 8.71 & 29.07 & 8.43  \\Qwen2.5-14B-Ins.    & \underline{85.00} & \underline{73.32} & \underline{48.30} & \underline{67.46} & \underline{46.18} & \underline{19.24} & \underline{39.47} & \underline{17.98}  \\
    Qwen3-8B            & 69.38 & 60.22 & 38.80 & 61.09 & 38.32 & 14.47 & 35.11 & 14.04  \\
    \midrule
    \multicolumn{9}{c}{\textit{Open-Sourced Models $>$ 20B}} \\
    \midrule
    LLaMA-3.1-70B-Ins.  & 90.49 & 49.28 & 20.12 & 45.59 & 19.75 & 3.51 & 18.12 & 3.37  \\
    Qwen3-32B           & 76.21 & 67.91 & 47.45 & 66.88 & 47.03 & 20.79 & 43.54 & 20.08 \\
    Qwen3-30B-A3B-Ins. & \underline{95.10} & \underline{82.17} & \underline{59.39} & \underline{76.70} & \underline{58.86} & 26.69 & \underline{52.67} & \underline{26.26}  \\
    DeepSeek-V3.1-Terminus    & 90.04 & 78.29 & 56.32 & 75.96 & 55.73 & \underline{26.97} & 52.39 & \underline{26.26} \\
    \bottomrule
    \end{tabularx}
    \vspace{-1.0 em}
\end{table*}

\begin{table*}[ht]\centering
    \caption{\textbf{Results of{\benchname}. All experiments are conducted with a delay of one to two turns randomized.}
    \label{table:result_time12}
    \textbf{Bold} indicates best overall performance, while \underline{underline} denotes the best within the same group. }
    \small
    \begin{tabularx}{\textwidth}{l|*{7}{>{\centering\arraybackslash}X}|>{\centering\arraybackslash}X}
    \toprule
    \multirow{2}{*}{\textbf{Models}}& \multicolumn{2}{c}{\textbf{Step Level}}
        & \multicolumn{3}{c}{\textbf{Sub-Task Level}} 
        & \multicolumn{2}{c|}{\textbf{Task Level}} 
        & \multirow{2}{*}{\textbf{Overall}} \\
    \cmidrule(lr){2-3} \cmidrule(lr){4-6} \cmidrule(lr){7-8}
        & Func. & Param. 
        & Char. & Env. & Acc. 
        & Char. & Env.
        &\\
    \midrule
    \multicolumn{9}{c}{\textit{Closed-Sourced Models}} \\
    \midrule
    Qwen-Max  & \underline{\textbf{96.87}} & \underline{\textbf{83.56}} & \underline{\textbf{64.49}} & \underline{\textbf{78.50}} & \underline{\textbf{61.94}} & \underline{\textbf{33.85}} & 54.63 & \underline{\textbf{30.20}} \\
    Gemini 2.5 Pro& 93.98 & 81.68 & 61.84 & 78.34 & 61.04 & 30.62 & \underline{\textbf{57.58}} & 29.35  \\
    GPT-4o    & 93.08 & 81.05 & 60.51 & 78.29 & 59.66 & 30.06 & 57.44 & 28.79 \\
    GPT-4.1   & 91.97 & 80.29 & 58.76 & 76.17 & 58.07 & 30.34 & 55.90 & 29.49 \\
    \midrule
    \multicolumn{9}{c}{\textit{Open-Sourced Models $<$ 20B}} \\
    \midrule
    LLaMA-3.1-8B-Ins.& 83.77 & 45.06 & 12.74 & 41.14 & 11.68 & 1.54 & 15.03 & 1.54  \\
    Qwen2.5-7B-Ins.     & 86.02 & 66.81 & 29.09 & 56.48 & 27.55 & 6.88 & 26.26 & 6.04 \\
    Qwen2.5-14B-Ins.    & \underline{87.22} & \underline{73.30} & \underline{49.10} & \underline{65.82} & \underline{45.70} & \underline{19.52} & \underline{36.94} & \underline{16.29} \\
    Qwen3-8B            & 60.72 & 50.76 & 25.85 & 51.38 & 25.42 & 7.30 & 25.14 & 7.16 \\
    \midrule
    \multicolumn{9}{c}{\textit{Open-Sourced Models $>$ 20B}} \\
    \midrule
    LLaMA-3.1-70B-Ins.  & 90.76 & 46.37 & 16.67 & 40.98 & 16.03 & 2.25 & 13.90 & 1.83 \\
    Qwen3-32B           & 71.41 & 61.92 & 37.10 & 59.55 & 36.20 & 12.08 & 33.71 & 11.24 \\
    Qwen3-30B-A3B-Ins. & \underline{95.39} & \underline{79.86} & \underline{54.09} & \underline{72.13} & \underline{52.87} & \underline{22.75} & \underline{46.49} & \underline{20.93} \\
    DeepSeek-V3.1-Terminus    & 84.52 & 73.42& 47.35 & 68.31 & 46.07  & 17.98 & 39.89 & 16.57 \\
    \bottomrule
    \end{tabularx}
    \vspace{-1.0 em}
\end{table*}

\begin{table*}[ht]
    \centering
    \caption{\textbf{Results of {\benchname} on fewshot settings. Specifically, we provide a successful trajectory in the prompt as a reference.} 
    \label{table:result_rebuttal_fewshot}}
    \small
    \begin{tabularx}{\textwidth}{l|*{7}{>{\centering\arraybackslash}X}|>{\centering\arraybackslash}X}
    \toprule
    \multirow{2}{*}{\textbf{Models}}
        & \multicolumn{2}{c}{\textbf{Step Level}} 
        & \multicolumn{3}{c}{\textbf{Sub-Task Level}} 
        & \multicolumn{2}{c|}{\textbf{Task Level}} 
        & \multirow{2}{*}{\textbf{Overall}} \\
    \cmidrule(lr){2-3} \cmidrule(lr){4-6} \cmidrule(lr){7-8}
        & Func. & Param. 
        & Char. & Env. & Acc. 
        & Char. & Env. 
        &  \\ 
    \midrule
    \multicolumn{9}{c}{\textit{Open-Sourced Models $<$ 20B}} \\
    \midrule
    Qwen2.5-7B-Ins.     
        & 82.40 & 65.01 & 28.66 & 55.15 & 26.38 & 6.74 & 25.84 & \textbf{6.04} \\
    \quad +fewshot      
        & 78.89 & 61.72 & 30.04 & 54.41 & 28.13 & 8.99 & 26.40 & \textbf{8.29} \\
    LLaMA-3.1-8B-Ins.   
        & 78.29 & 43.69 & 12.95 & 41.08 & 12.47 & 1.26 & 14.61 & \textbf{1.26} \\
    \quad +fewshot      
        & 90.27 & 63.35 & 29.67 & 56.37 & 28.03 & 7.16 & 27.67 & \textbf{6.74} \\
    Qwen3-8B            
        & 63.05 & 53.61 & 29.88 & 53.08 & 29.30 & 11.10 & 28.65 & \textbf{10.67} \\
    \quad +fewshot      
        & 72.04 & 60.67 & 34.82 & 34.82 & 34.34 & 11.52 & 32.30 & \textbf{11.24} \\
    Qwen2.5-14B-Ins.    & 81.32 & 70.22 & 50.21 & 65.29 & 46.28 & 22.61 & 38.20 & \textbf{18.32} \\
    \quad +fewshot      & 94.28 & 80.40 & 54.19 & 75.32 & 52.87 & 22.89 & 50.14 & \textbf{21.91} \\
    Qwen2.5-72B-Ins.
        & 95.69 & 82.78 & 65.07 & 79.78 & 62.00 & 34.41 & 58.01 & \textbf{31.04} \\
    \quad +fewshot
        & 89.34 & 83.61 & 66.40 & 81.16 & 64.65 & 36.94 & 59.69 & \textbf{34.55} \\
    \bottomrule
    \end{tabularx}
    \vspace{-1.0 em}
\end{table*}

\begin{table*}[t]
    \centering
    \caption{\textbf{Results of {\benchname}. Specifically, we analyze the average number of rounds of interaction in successfully completed task models.}}
    \label{table:result_avg_turns}
    \small
    \begin{tabularx}{0.7\textwidth}{l|>{\centering\arraybackslash}X|>{\centering\arraybackslash}X}
    \toprule
    \textbf{Models} & \textbf{\# 2} & \textbf{\# 3} \\
    \midrule
    \multicolumn{3}{c}{\textit{\textbf{Closed-Sourced Models}}} \\
    \midrule
    Qwen-max & 6.91 & 9.38 \\
    Gemini 2.5 Pro & 6.85 & 10.50 \\
    GPT-4o & 6.95 & 10.52 \\
    GPT-4.1 & 6.70 & 10.57 \\
    \midrule
    \multicolumn{3}{c}{\textit{\textbf{Open-Sourced Models $<$ 20B}}} \\
    \midrule
    LLaMA-3.1-8B-Ins. & 9.75 & 9.00 \\
    Qwen2.5-7B-Ins. & 8.50 & 12.35 \\
    Qwen2.5-14B-Ins. & 7.94 & 10.95 \\
    Qwen3-8B & 6.64 & 9.19 \\
    \midrule
    \multicolumn{3}{c}{\textit{\textbf{Open-Sourced Models $>$20B}}} \\
    \midrule
    LLaMA-3.1-70B-Ins. & 8.41 & 16.28 \\
    LLaMA-3.3-70B-Ins. & 10.62 & 16.23 \\
    Qwen2.5-32B-Ins. & 9.20 & 13.68 \\
    \bottomrule
    \end{tabularx}
\end{table*}

\label{app:analysis}

\clearpage
\onecolumn

\section{Use of Large Language Models}
We use a large language model for translation and language polishing, but its role in the paper is limited. The core research and content are completed by us.

\definecolor{lightgray}{gray}{0.95}
\definecolor{deepblue}{RGB}{70,130,180}
\definecolor{deepgray}{RGB}{119,136,153}
\definecolor{PeachPuff3}{RGB}{205,175,149}
\definecolor{RosyBrown}{RGB}{188,143,143}
\lstdefinestyle{prompt}{
    basicstyle=\ttfamily\fontsize{7pt}{8pt}\selectfont,
    frame=none,
    breaklines=true,
    backgroundcolor=\color{lightgray},
    breakatwhitespace=true,
    breakindent=0pt,
    escapeinside={(*@}{@*)},
    numbers=none,
    numbersep=5pt,
    xleftmargin=5pt,
    aboveskip=2pt,
    belowskip=2pt,
}
\tcbset{
  aibox/.style={
    top=10pt,
    colback=white,
    enhanced,
    breakable,
    width=\linewidth,
    halign=flush left,
  }
}

\newtcolorbox{AIbox}[2][]{aibox, title=#2,#1}

\newtcblisting{ErrorCodeBox}[1][]{%
  aibox,
  enhanced,
  listing only,
  listing options={style=prompt,#1},
  overlay={\node[anchor=north east, text=red, font=\Large, xshift=-2pt, yshift=-2pt] 
    at (frame.north east) {$\times$};},
}

\begin{AIbox}{Standardization of Test Data}
\textbf{System Prompt:} \\
You are an expert in handling multiple tasks using functions. Your goal is to call one function in each round, gradually completing all tasks. When a function call is not yet complete, you can switch to another task.
If you decide to invoke a function for a task, you MUST format it as follows: 
\begin{lstlisting}[style=prompt]
{"id": "id of task", "func_name": "func_name", "params": {"params_name1": "params_value1", "params_name2": "params_value2"}} 
\end{lstlisting}
You must invoke only one function in your response.
If you believe all tasks have been completed, please use the following format to reply:
\begin{lstlisting}[style=prompt]
{"content": "ALL COMPLETED"}
\end{lstlisting}
Here is a list of functions in JSON format that you can invoke.
\begin{lstlisting}[style=prompt]
[API List]
\end{lstlisting}
\tcblower
{\color{deepblue}\bf User:}
Below are all tasks you need to finish:
\begin{lstlisting}[style=prompt]
{
    id:'file_13',
    question:'Compare 'academic_venture/goals.txt' with 'academic_venture/reference_goals.txt' to verify if they are identical, then rename 'goals.txt' to 'master_goals.txt' in the same directory
    tory.'
}
{
    id:'SM_11',
    question have a list of integers [1, 2, 3, 4, 5]. I want to convert this list into a string representing a space-separated sequence of those integers. Then, I want to normalize this string by replacing all non-alphanumeric characters with underscores and converting to lowercase.'
}
{
    id:'posting_11',
    question:'Now that I'm following 'AI_enthusiast', I'd like to know more about them
    . Please retrieve their user statistics. After that, fetch all of their tweets. I remember seeing one about 'ethics'; find that specific tweet and add the comment 'A very important point to consider.' to it.'
}
\end{lstlisting}
{\color{teal}\bf Ground True:}
\begin{lstlisting}[style=prompt]
[{
    "task_id": "file_13",
    "ground_truth": [
          "cd(folder='academic_venture')",
          "diff(file_name1='goals.txt', file_name2='reference_goals.txt')",
          "mv(source='goals.txt', destination='master_goals.txt')"
        ],
    },
    {
    "task_id": "SM_11",
    "ground_truth": [
          "integer_list_to_string(integer_list=[1, 2, 3, 4, 5])",
          "normalize_string(s='1 2 3 4 5')"
        ],
    },
    {
    "task_id": "posting_11",
    "ground_truth": [
          "get_user_stats(username='AI_enthusiast')",
          "get_user_tweets(username='AI_enthusiast')",
          "comment(tweet_id=18, comment_content='A very important point to consider.')"
        ],
}]
\end{lstlisting}
\end{AIbox}

\captionof{figure}{An example of standardized test data.}
\label{fig:test-data-standardization}

\clearpage
\begin{AIbox}{Prompt for Reconstruction}
\textbf{Prompt:} \\
You are an AI assistant with API-calling capabilities. Your task is to refine query descriptions based on given contexts and initial configurations, then provide explicit solution paths. You must meet these requirements:
\begin{enumerate}[leftmargin=*]
\item Ensure problems are solvable within the given initial configuration, avoiding ambiguous or unsolvable cases. Clarify any ambiguities in the original problem by deriving explicit specifications from the initial configuration. Including but not limited to: file names, folder paths, numerical counts, and similar parameters.
\item Define solution paths as strict API call sequences (not less than 2 steps) with strong dependencies—later steps must require outputs from earlier ones. Expand the problem if necessary to meet this length requirement.
\item Preserve existing explicit information (\eg, filenames, data). Extend problems only by deriving new details from the initial config.
\item Always provide the most efficient path—no unnecessary API calls. Consecutive duplicate function calls are prohibited in the execution sequence.
\item Never explicitly name APIs in the problem description. Carefully review the function description. You must ensure all function calls in the provided ground truth strictly comply with the specifications. For example, certain filesystem operations are restricted to the current working directory and cannot process multi-level paths. Only functions explicitly defined in the function description may appear in the ground truth.
6. For every input data given, you must return the same number of corresponding processed data.
\item You have permission to edit the question and ground\_truth fields, and your response must explicitly return modified versions. The output format should follow the usage examples in \textbf{JSON format} instead of the origin data.
\end{enumerate}
You can use the following APIs:
\begin{lstlisting}[style=prompt]
[API Document]
\end{lstlisting}
Here is one usage examples that you must follow:
\begin{lstlisting}[style=prompt]
[
    {
        "id": "number_0", 
        "question":"Hey there, I need to set up a directory titled 'Reports' in my current workspace. Once that's in place, could you help me locate a file called 'summary.doc' in this vicinity and transfer it to the new 'Reports' folder? After that, I'd appreciate if you could find and open up 'data.txt', then scan through it to identify lines that mention 'Q4 financials'. Also, could you let me know the total number of lines in 'data.txt'?",
        "ground_truth":["mkdir(dir_name='Reports')","mv(source='summary.doc', destination='Reports')", "cat(file_name='data.txt')", "grep(pattern='Q4 financials', file_name='data.txt')", "wc(file_name='data.txt',mode='l')"],
        "thought":"your thought"
    },
    {
        "id": "number_1",
        "question":"Let's bring some order to the project documents. I want to human-readable log the storage usage of the entire current directory to usage.txt file. The content of the file should be the number follwed by the word bytes and nothing else.",
        "ground_truth":["du(human_readable=True)", "touch(file_name='usage.txt')", "echo(content='205 bytes', file_name='usage.txt')"],
        "thought":"your thought"
    }
]
\end{lstlisting}
where `id' is the id of task, `question' is the user query, `ground\_truth' is the shortest function calls sequences used to solve the query, and `thought' is your ideas for refining this query.
\end{AIbox}

\captionof{figure}{prompt for reconstruction.}
\label{fig:test-data-reconstruction-prompt}

\clearpage
\begin{AIbox}{Reconstructed Data}
\textbf{Task:} \\
Locate a file with ``ideas'' in its name within the `documents' folder, display its contents, and create a backup copy named `ideas\_backup.txt' in the same directory. \\
\textbf{Config:} \\
\begin{lstlisting}[style=prompt]
{
  "root": {
    "simona": {
      "type": "directory",
      "contents": {
        "documents": {
          "type": "directory",
          "contents": {
            "ideas.txt": {
              "type": "file",
              "content": "Collaboration leads to success. Innovation ignites growth."
            },
            "Archived": {
              "type": "directory",
              "contents": {}
            },
            "past_projects": {
              "type": "directory",
              "contents": {}
            }
          }
        }
      }
    }
  }
}
\end{lstlisting}
\textbf{Trajectory:} \\
\begin{lstlisting}[style=prompt]
[
  "cd(folder='documents')",
  "find(name='ideas')",
  "cat(file_name='ideas.txt')",
  "cp(source='ideas.txt', destination='ideas_backup.txt')"
]
\end{lstlisting}
\end{AIbox}

\captionof{figure}{An example of reconstructed data: GorillaFileSystem.}
\label{fig:test-data-reconstructed-data}

\clearpage
\begin{AIbox}{API Document}
\begin{lstlisting}[style=prompt]
{
    "name": "mv",
    "description": "This tool belongs to the Gorilla file system. It is a simple file system that allows users to perform basic file operations such as navigating directories, creating files and directories, reading and writing to files, etc. Tool description: Move a file or directory from one location to another. so",
    "parameters": {
        "type": "dict",
        "properties": {
            "source": {
                "type": "string",
                "description": "Source name of the file or directory to move. Source must be local to the current directory."
            },
            "destination": {
                "type": "string",
                "description": "The destination name to move the file or directory to. Destination must be local to the current directory and cannot be a path. If destination is not an existing directory like when renaming something, destination is the new file name. "
            }
        },
        "required": [
            "source",
            "destination"
        ]
    },
    "response": {
        "type": "dict",
        "properties": {
            "result": {
                "type": "string",
                "description": "The result of the move operation."
            }
        }
    }
}
\end{lstlisting}
\end{AIbox}

\captionof{figure}{An example of an API document.}
\label{fig:test-data-api-document}

\clearpage
\begin{AIbox}{Correct Trajectory}
\textbf{User:} \\
Below are all tasks you need to finish:
\begin{lstlisting}[style=prompt]
{id:'trading_0',question:'Can you get me the ticker for 'Alpha Tech', see its current price, and then execute a purchase of 20 shares at a price of $1320.5?'}
{id:'file_11',question:'Hey, can you set up a new directory named 'Projects' right in workspace folder?'}
\end{lstlisting}
\tcblower
{\color{RosyBrown}\bf Assistant:}
\begin{lstlisting}[style=prompt]
{
    "id": "trading_0", 
    "func_name": "get_symbol_by_name", 
    "params": {"name": "Alpha Tech"}
}
\end{lstlisting}
{\color{PeachPuff3}\bf Environment:}
\begin{lstlisting}[style=prompt]
{"role": "tool", "content": "{"wait": "The current function is being executed."}"}
\end{lstlisting}
{\color{RosyBrown}\bf Assistant:}
\begin{lstlisting}[style=prompt]
{
    "id": "file_11", 
    "func_name": "cd", 
    "params": {"folder": "workspace"}
}
\end{lstlisting}
{\color{PeachPuff3}\bf Environment:}
\begin{lstlisting}[style=prompt]
{"role": "tool", "content": "{"wait": "The current function is being executed."}"},
{"role": "tool", "id": "trading_0", "function name": "get_symbol_by_name(name="Alpha Tech")", "response": "{"symbol": "ALPH"}"}
\end{lstlisting}
{\color{RosyBrown}\bf Assistant:}
\begin{lstlisting}[style=prompt]
{
    "id": "trading_0", 
    "func_name": "get_stock_info", 
    "params": {"symbol": "ALPH"}
}
\end{lstlisting}
{\color{PeachPuff3}\bf Environment:}
\begin{lstlisting}[style=prompt]
{"role": "tool", "content": "{"wait": "The current function is being executed."}"},
{"role": "tool", "id": "file_11", "function name": "cd(folder="workspace")", "response": "{"current_working_directory": "workspace"}"}
\end{lstlisting}
{\color{RosyBrown}\bf Assistant:}
\begin{lstlisting}[style=prompt]
{
    "id": "file_11", 
    "func_name": "mkdir", 
    "params": {"dir_name": "Projects"}
}
\end{lstlisting}
{\color{PeachPuff3}\bf Environment:}
\begin{lstlisting}[style=prompt]
{"role": "tool", "content": "{"wait": "The current function is being executed."}"},
{"role": "tool", "id": "trading_0", "function name": "get_stock_info(symbol="ALPH")", "response": "{"price": 1320.45, "percent_change": -0.08, "volume": 1.567, "MA(5)": 1321.12, "MA(20)": 1325.78}"}
\end{lstlisting}
{\color{RosyBrown}\bf Assistant:}
\begin{lstlisting}[style=prompt]
{
    "id": "trading_0", 
    "func_name": "place_order", 
    "params": {"order_type": "Buy", "symbol": "ALPH", "price": 1320.5, "amount": 20}
}
\end{lstlisting}
{\color{PeachPuff3}\bf Environment:}
\begin{lstlisting}[style=prompt]
{"role": "tool", "content": "{"wait": "The current function is being executed."}"},
{"role": "tool", "id": "file_11", "function name": "mkdir(dir_name="Projects")", "response": "None"}
\end{lstlisting}
{\color{RosyBrown}\bf Assistant:}
\begin{lstlisting}[style=prompt]
{"content": "ALL COMPLETED"}
\end{lstlisting}
\end{AIbox}

\captionof{figure}{An example of correct trajectory.}
\label{fig:test-data-correct-trajectory}

\clearpage
\begin{AIbox}{Error Caused by Insufficient Temporal Awareness}
\textbf{User:} \\
Below are all tasks you need to finish:
\begin{lstlisting}[style=prompt]
{
    id:'trading_0',
    question:'Can you get me the ticker for 'Alpha Tech', see its current price, and then execute a purchase of 20 shares at a price of $1320.5?'
}
{
    id:'file_11',
    question:'Hey, can you set up a new directory named 'Projects' right in workspace folder?'
}
\end{lstlisting}
\tcblower
{\color{RosyBrown}\bf Assistant:}
\begin{lstlisting}[style=prompt]
{
    "id": "trading_0", "func_name": "get_symbol_by_name",
    "params": {"name": "Alpha Tech"}
}
\end{lstlisting}
{\color{PeachPuff3}\bf Environment:}
\begin{lstlisting}[style=prompt]
{
    "role": "tool",
    "content": "{"wait": "The current function is being executed."}"
}
\end{lstlisting}
{\color{RosyBrown}\bf Assistant:}
\begin{lstlisting}[style=prompt,
literate={ATGL}{{{\color{red}ATGL}}}4]
{
    "id": "trading_0", "func_name": "get_stock_info",
    "params": {"symbol": "ATGL"}
}
\end{lstlisting}
\end{AIbox}

\captionof{figure}{An example of an error caused by insufficient temporal awareness. In this case, the agent prematurely assumed the symbol of Alpha Tech to be ``ATGL'' before receiving the actual call result.}
\label{fig:test-data-temporal-error}

\clearpage
\begin{AIbox}{Error Caused by Tool Confusion}
\textbf{User:} \\
Below are all tasks you need to finish:
\begin{lstlisting}[style=prompt]
{
    id:'trading_0',
    question:'Can you get me the ticker for 'Alpha Tech', see its current price, and then execute a purchase of 20 shares at a price of $1320.5?'
}
{
    id:'file_11',
    question:'Hey, can you set up a new directory named 'Projects' right in workspace folder?'
}
\end{lstlisting}
\tcblower
{\color{RosyBrown}\bf Assistant:}
\begin{lstlisting}[style=prompt]
{
    "id": "trading_0", "func_name": "get_symbol_by_name",
    "params": {"name": "Alpha Tech"}
}
\end{lstlisting}
{\color{PeachPuff3}\bf Environment:}
\begin{lstlisting}[style=prompt]
{
    "role": "tool",
    "content": "{"wait": "The current function is being executed."}"
}
\end{lstlisting}
{\color{RosyBrown}\bf Assistant:}
\begin{lstlisting}[style=prompt,
literate={trading_0}{{{\color{red}trading\_0}}}9]
{
    "id": "trading_0", "func_name": "cd",
    "params": {"folder": "workspace"}
}
\end{lstlisting}
\end{AIbox}

\captionof{figure}{An example of an error caused by tool confusion. In this case, the agent mistakenly applied the tool intended for task ``file\_11'' to task ``trading\_0''.}
\label{fig:test-data-tool-confusion}

\clearpage
\begin{AIbox}{Error Caused by Neglecting a Certain Task}
\textbf{User:} \\
Below are all tasks you need to finish:
\begin{lstlisting}[style=prompt]
{
    id:'file_13',
    question:'Compare 'academic_venture/goals.txt' with 'academic_venture/reference_goals.txt' to verify if they are identical, then rename 'goals.txt' to 'master_goals.txt' in the same directory.'
}
{
    id:'SM_11',
    question:'I have a list of integers [1, 2, 3, 4, 5]. I want to convert this list into a string representing a space-separated sequence of those integers. Then, I want to normalize this string by replacing all non-alphanumeric characters with underscores and converting to lowercase.'
}
{
    id:'posting_11',
    question:'Now that I'm following 'AI_enthusiast', I'd like to know more about them. Please retrieve their user statistics. After that, fetch all of their tweets. I remember seeing one about 'ethics'; find that specific tweet and add the comment 'A very important point to consider.' to it.'
}
\end{lstlisting}
\tcblower
{\color{RosyBrown}\bf Assistant:}
\begin{lstlisting}[style=prompt]
{"id": "file_13", "func_name": "cd", "params": {"folder": "academic_venture"}}
\end{lstlisting}
{\color{PeachPuff3}\bf Environment:}
\begin{lstlisting}[style=prompt]
{"role": "tool", "content": "{"wait": "The current function is being executed."}"}
\end{lstlisting}
{\color{RosyBrown}\bf Assistant:}
\begin{lstlisting}[style=prompt]
{"id": "SM_11", "func_name": "integer_list_to_string", "params": {"integer_list": [1, 2, 3, 4, 5]}}
\end{lstlisting}
{\color{PeachPuff3}\bf Environment:}
\begin{lstlisting}[style=prompt]
{"role": "tool", "content": "{"wait": "The current function is being executed."}"},{"role": "tool", "id": "file_13", "function name": "cd(folder="academic_venture")", "response": "{"current_working_directory": "academic_venture"}"}
\end{lstlisting}
{\color{RosyBrown}\bf Assistant:}
\begin{lstlisting}[style=prompt]
{"id": "file_13", "func_name": "diff", "params": {"file_name1": "goals.txt", "file_name2": "reference_goals.txt"}}
\end{lstlisting}
{\color{PeachPuff3}\bf Environment:}
\begin{lstlisting}[style=prompt]
{"role": "tool", "content": "{"wait": "The current function is being executed."}"},{"role": "tool", "id": "SM_11", "function name": "integer_list_to_string(integer_list=[1, 2, 3, 4, 5])", "response": "1 2 3 4 5"}
\end{lstlisting}
{\color{RosyBrown}\bf Assistant:}
\begin{lstlisting}[style=prompt]
{"id": "SM_11", "func_name": "normalize_string", "params": {"s": "1 2 3 4 5"}}
\end{lstlisting}
{\color{PeachPuff3}\bf Environment:}
\begin{lstlisting}[style=prompt]
{"role": "tool", "content": "{"wait": "The current function is being executed."}"},{"role": "tool", "id": "file_13", "function name": "diff(file_name1="goals.txt", file_name2="reference_goals.txt")", "response": "{"diff_lines": "- Research topic selection\\\\n+ Data analysis\\\\n- Literature review\\\\n+ Data collection\\\\n- Data collection\\\\n+ Draft writing\\\\n- Data analysis\\\\n+ Final submission\\\\n- Draft writing\\\\n+ Literature review\\\\n- Final submission\\\\n+ Research topic selection"}"}
\end{lstlisting}
{\color{RosyBrown}\bf Assistant:}
\begin{lstlisting}[style=prompt]
{"id": "file_13", "func_name": "mv", "params": {"source": "goals.txt", "destination": "master_goals.txt"}}
\end{lstlisting}
{\color{PeachPuff3}\bf Environment:}
\begin{lstlisting}[style=prompt]
{"role": "tool", "content": "{"wait": "The current function is being executed."}"},{"role": "tool", "id": "SM_11", "function name": "normalize_string(s="1 2 3 4 5")", "response": "1_2_3_4_5"}
\end{lstlisting}
{\color{RosyBrown}\bf Assistant:}
\begin{lstlisting}[style=prompt]
{"content": "ALL COMPLETED"}
\end{lstlisting}
\end{AIbox}

\captionof{figure}{An example of an error caused by neglecting a certain task. In this case, the task ``posting\_11'' was ignored by the agent.}
\label{fig:test-data-neglect-task}

\end{document}